\documentclass[journal]{IEEEtran}

\usepackage{amsmath,amsfonts}
\usepackage{algorithmic}
\usepackage{array}
\usepackage[caption=false,font=normalsize,labelfont=sf,textfont=sf]{subfig}
\usepackage{textcomp}
\usepackage{stfloats}
\usepackage{url}
\usepackage{hyperref}
\usepackage{verbatim}
\usepackage{graphicx}
\usepackage{xcolor,colortbl}
\usepackage{tabularx,booktabs}
\usepackage{array,multirow,graphicx}
\usepackage{makecell}
\usepackage{amssymb}
\usepackage{wasysym}
\usepackage[export]{adjustbox}
\usepackage{pifont} 
\usepackage{caption}
\usepackage{etoolbox}
\usepackage[normalem]{ulem}

\definecolor{tabfirst}{rgb}{1, 0.7, 0.7} 
\definecolor{tabsecond}{rgb}{1, 0.85, 0.7} 
\definecolor{tabthird}{rgb}{1, 1, 0.7} 
\newcommand{\fr}{{\cellcolor{tabfirst}}}
\newcommand{\nd}{{\cellcolor{tabsecond}}}
\newcommand{\rd}{{\cellcolor{tabthird}}}

\newcommand{\ml}[2]{\multicolumn{#1}{l}{#2}}
\definecolor{LightCyan}{rgb}{0.88,0.88,0.88}
\DeclareMathOperator*{\argminA}{arg\,min} 

\newif\ifrevision
\revisionfalse
\newcommand{\revadd}[1]{\ifrevision{\color{blue}{#1}}\else{#1}\fi}
\newcommand{\revdel}[1]{\ifrevision{\color{red}{\sout{#1}}}\else{}\fi}

\def\BibTeX{{\rm B\kern-.05em{\sc i\kern-.025em b}\kern-.08em
    T\kern-.1667em\lower.7ex\hbox{E}\kern-.125emX}}
\usepackage{balance}

\begin{document}
\title{HI-SLAM2: Geometry-Aware Gaussian SLAM for Fast Monocular Scene Reconstruction}
\author{Wei Zhang$^{1}$, Qing Cheng$^{2,4}$, David Skuddis$^{1}$, Niclas Zeller$^{3}$, Daniel Cremers$^{2,4}$ and Norbert Haala$^{1}$
\thanks{$^{1}$Institute for Photogrammetry and Geoinformatics, University of Stuttgart, Germany
        {\tt\small \{firstname.lastname\}@ifp.uni-stuttgart.de}}%
\thanks{$^{2}$Technical University of Munich, Germany}%
\thanks{$^{3}$Karlsruhe University of Applied Sciences, Germany}%
\thanks{$^{4}$Munich Center for Machine Learning, Germany}%
}

\markboth{Published in IEEE Transactions on Robotics. DOI: \href{https://doi.org/10.1109/TRO.2025.3626627}{10.1109/TRO.2025.3626627}}
{How to Use the IEEEtran \LaTeX \ Templates}

\newcommand{\insertfig}{
   \vspace{1.3em}
   \includegraphics[width=\textwidth]{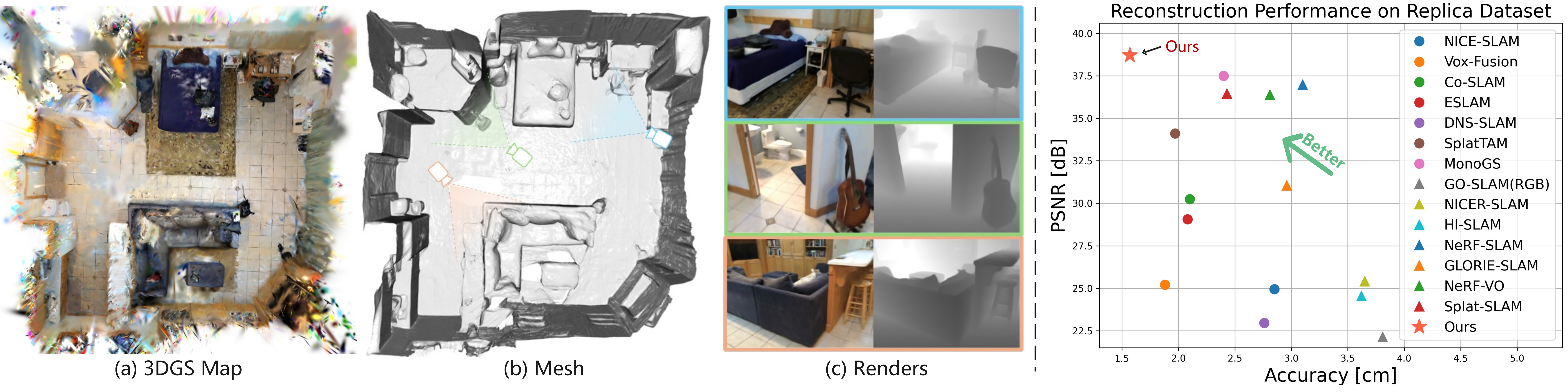}  
   \vspace{-1.7em}
   \setcounter{figure}{0}
   \captionof{figure}{Our method builds a 3D Gaussian Splatting (3DGS) map (a) to reconstruct complex scenes using only monocular input. We are able to extract accurate and detailed mesh reconstructions (b) with high-quality renderings (c). The right figure illustrates the trade-off between geometric accuracy and visual appearance, as some methods prioritize one aspect over the other. Compared to existing methods, our approach excels among RGB-only methods, denoted by $\blacktriangle$, and also surpasses recent RGB-D methods, denoted by $\CIRCLE$, in both geometry and appearance reconstruction.}
  \label{fig:teaser}
  \vspace{-2.2em}
}

\makeatletter
\apptocmd{\@maketitle}{\insertfig}{}{}
\makeatother

\maketitle

\begin{abstract}
We present HI-SLAM2, a geometry-aware Gaussian SLAM system that achieves fast and accurate monocular scene reconstruction using only RGB input. Existing Neural SLAM or 3DGS-based SLAM methods often trade off between rendering quality and geometry accuracy, our research demonstrates that both can be achieved simultaneously with RGB input alone. The key idea of our approach is to enhance the ability for geometry estimation by combining easy-to-obtain monocular priors with learning-based dense SLAM, and then using 3D Gaussian splatting as our core map representation to efficiently model the scene. Upon loop closure, our method ensures on-the-fly global consistency through efficient pose graph bundle adjustment and instant map updates by explicitly deforming the 3D Gaussian units based on anchored keyframe updates. Furthermore, we introduce a grid-based scale alignment strategy to maintain improved scale consistency in prior depths for finer depth details. Through extensive experiments on Replica, ScanNet, Waymo Open, ETH3D SLAM and ScanNet++ datasets, we demonstrate significant improvements over existing Neural SLAM methods and even surpass RGB-D-based methods in both reconstruction and rendering quality. The project page and source code \revdel{will be made}\revadd{are} available at \href{https://hi-slam2.github.io/}{https://hi-slam2.github.io/}.
\end{abstract}


\section{Introduction}
\IEEEPARstart{D}{ense} 3D scene reconstruction from imagery remains one of the most fundamental challenges in computer vision, robotics, and photogrammetry. Achieving real-time and accurate 3D reconstruction from images alone can enable numerous applications, from autonomous navigation to mobile surveying and immersive AR. While many existing solutions rely on RGB-D~\cite{newcombe2011kinectfusion, whelan2015elasticfusion, dai2017bundlefusion, azinovic2022neural, sucar2021imap} or LiDAR sensors~\cite{lu1997globally, zhang2014loam, hess2016real, xu2022fast, pan2024pin}, these approaches have inherent limitations. LiDAR systems require expensive hardware setups and an additional camera for capturing color information, while RGB-D sensors suffer from limited operational range and sensitive to varying lighting conditions. Vision-based monocular scene reconstruction thus offers a promising lightweight and cost-effective alternative.

The fundamental challenge in monocular 3D reconstruction stems from the lack of explicit scene geometry measurements~\cite{cadena2016past}. Traditional visual SLAM methods~\cite{davison2007monoslam, mur2015orb, engel2014lsd, forster2016svo, qin2018vins, campos2021orb} developed over decades and typically provide only sparse or semi-dense map representations, proving insufficient for detailed scene understanding and complete reconstruction. While dense SLAM approaches~\cite{newcombe2011dtam, engel2017direct, teed2021droid} attempt to address this limitation through per-pixel depth estimation, they remain susceptible to significant depth noise and struggle to achieve complete, accurate reconstructions.\par

Recent advances in deep learning have revolutionized many key components of 3D reconstruction, including optical flow~\cite{teed2020raft, xu2022gmflow}, depth estimation~\cite{yao2018mvsnet, eftekhar2021omnidata}, and normal estimation~\cite{eftekhar2021omnidata, bae2021estimating}. These improvements have been integrated into SLAM systems through monocular depth networks~\cite{tateno2017cnn}, multi-view stereo techniques~\cite{teed2018deepv2d}, and end-to-end neural approaches~\cite{godard2019digging}. However, even with these advancements, current systems often produce reconstructions with artifacts due to noisy depth estimates, limited generalization capability, or excessive computational requirements. The emergence of Neural SLAM methods, particularly those based on neural implicit fields~\cite{sucar2021imap, zhu2022nice, rosinol2022nerf} and 3D Gaussian Splatting (3DGS)~\cite{keetha2024splatam, matsuki2024gaussian, sandstrom2024splat}, has shown promising results. Yet these approaches typically prioritize either rendering quality or geometry accuracy, creating an undesirable trade-off. Our work addresses this limitation by simultaneously improving both aspects without compromise either. As shown in Fig.~\ref{fig:teaser}, our approach achieves superior performance across both geometry accuracy and rendering quality, surpassing not only RGB-based methods but also RGB-D-based approaches.\par

\begin{figure}[t!]
\centering
\includegraphics[width=0.496\textwidth]{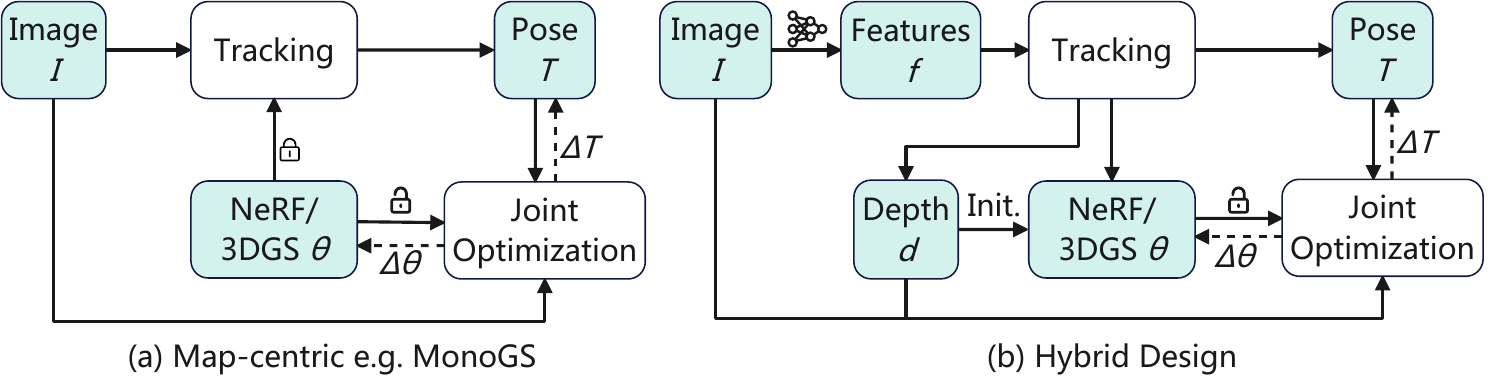}
\vspace*{-5mm}
\caption{Comparison of SLAM paradigms: while map-centric SLAM employs a unified map representation for both tracking and joint optimization, the hybrid design approach utilizes learning-based features and bundle adjustment for tracking, producing depth as an intermediate scene representation. This is then used to initialize the 3D map and supervise the joint optimization of camera poses and scene geometry.}
\label{fig:slam_paradigm}
\end{figure}

In this paper, we aim to advance the state-of-the-art in dense monocular SLAM for 3D scene reconstruction. We present HI-SLAM2, a geometry-aware Gaussian Splatting SLAM system that achieves accurate and fast monocular scene reconstruction using RGB input alone. The key idea of our approach lies in enhancing geometry estimation by combining monocular geometry priors with learning-based dense SLAM, while leveraging 3DGS as our compact map representation for efficient and accurate scene modeling. As depicted in Fig.~\ref{fig:slam_paradigm}, unlike map-centric SLAM methods, we adopt a hybrid approach that utilizes learning-based dense SLAM to generate depth as a proxy, which serves both to initialize scene geometry and to guide map optimization. This hybrid design decouples the map training from tracking while seamlessly recoupling pose and map later during joint optimization, ensuring both efficiency and accuracy.

For depth estimation, we introduce a scale-grid based alignment strategy that effectively addresses scale distortions in monocular depth priors, significantly improving depth estimation accuracy. Our surface depth rendering employs unbiased depth calculation at ray-Gaussian intersection points~\cite{zhang2024rade}, enabling more precise surface fitting. To enhance surface reconstruction, particularly in low-texture regions, we incorporate monocular normal priors into 3DGS training, ensuring the consistency of reconstructed surfaces. By deforming 3D Gaussian units using keyframe pose updates, we enable efficient online map updates, boosting both speed and flexibility in mapping. Furthermore, unlike hash grid based methods~\cite{wang2023co,zhang2024hi} that require a predefined scene boundary, our approach allows the map to grow incrementally as new areas are explored without any prior knowledge of scene size.

We validate our approach through extensive experiments on both synthetic and real-world datasets, including Replica~\cite{straub2019replica}, ScanNet~\cite{dai2017scannet}, Waymo Open~\cite{sun2020scalability}, ETH3D SLAM~\cite{Schops_2019_CVPR}, and ScanNet++~\cite{yeshwanth2023scannet++}. Our method achieves substantial improvements in both reconstruction and rendering quality compared to existing Neural SLAM methods, surpassing even RGB-D-based methods in accuracy. Our method is particularly well-suited for real-time applications that demand rapid and reliable scene reconstruction in scenarios where depth sensors are impractical.

In summary, our work advances the state-of-the-art in dense monocular SLAM through the following contributions:

\begin{itemize}
   \item A geometry-aware Gaussian SLAM framework achieving high-fidelity RGB-only reconstruction through efficient online mapping and joint optimization of camera poses and Gaussian map.

   \item An enhanced depth estimation approach leveraging geometry priors and improved scale alignment to compensate for monocular prior distortions and enable accurate surface reconstruction.

   \item A balanced system achieving superior performance in both geometry and appearance reconstruction across synthetic and real-world datasets.
\end{itemize}

\section{Related Works}
\subsection{Depth Estimation}
Depth estimation can be broadly categorized into multi-view and monocular approaches. Classic multi-view methods rely on geometry principles, utilizing techniques such as patch matching~\cite{schonberger2016structure} or cost aggregation~\cite{hirschmuller2007stereo}. Recent learning-based approaches MVSNet~\cite{yao2018mvsnet} and DeepMVS~\cite{huang2018deepmvs} have greatly improved the consistency of depth estimation across video sequences. In parallel, monocular depth estimation has seen remarkable progress, with methods like MiDaS~\cite{Ranftl2022} and OmniData~\cite{eftekhar2021omnidata} demonstrating impressive generalization across diverse datasets. However, these monocular approaches suffer from scale ambiguity, producing depth maps with inconsistent scales between frames. Our work addresses this limitation through a novel scale-grid alignment strategy that estimates spatially varying depth scales, enabling more accurate depth estimation compared to previous method~\cite{zhang2024hi} that relied on a single, rigid scale transformation.

\subsection{Surface Reconstruction}
Surface reconstruction typically follows a two-stage pipeline: camera pose estimation through Structure-from-Motion (SfM)~\cite{wu2011visualsfm, schonberger2016structure}, followed by multi-view stereo~\cite{hirschmuller2005accurate, barnes2009patchmatch} for dense reconstruction. While widely adopted, these methods are computationally intensive and often produce incomplete reconstructions due to depth estimation uncertainties. Neural implicit representations~\cite{mildenhall2021nerf} and their variants~\cite{muller2022instant, chen2022tensorf, fridovich2022plenoxels} have demonstrated high-quality reconstruction capabilities but remain computationally demanding. Recent advances in 3DGS~\cite{kerbl20233d} offer more efficient rendering compared to NeRF-based approaches, and its variants~\cite{huang20242d, zhang2024rade} show promising geometry reconstruction capabilities. Our approach leverages 3DGS for efficient scene representation while maintaining high-quality reconstruction, effectively addressing the speed-quality trade-off inherent in previous methods.

\subsection{Dense Visual SLAM}
Dense SLAM methods traditionally relied on volumetric representations such as Truncated Signed Distance Functions (TSDF)~\cite{curless1996volumetric} or 3D voxel grids~\cite{koestler2022tandem, zuo2023incremental} for scene geometry modeling. The emergence of neural implicit representations~\cite{mildenhall2021nerf} enabled high-quality scene reconstructions within dense visual SLAM~\cite{sucar2021imap, zhu2022nice}, but at significant computational cost and often requiring RGB-D input. Recent monocular approaches like NICER-SLAM~\cite{zhu2023nicer} and HI-SLAM~\cite{zhang2024hi} have demonstrated promising results using only RGB input. The 3DGS-based methods Splat-SLAM~\cite{sandstrom2024splat} and GLORIE-SLAM~\cite{zhang2024glorie} showcase the potential of 3DGS for real-time dense reconstruction. However, these methods still face challenges in balancing computational efficiency with reconstruction quality. Our work addresses these limitations through key innovations in depth estimation, scene consistency, and computational efficiency, achieving both high-quality geometry and appearance reconstruction in real-time.

\begin{figure*}[t!]
\centering
\includegraphics[width=0.95\textwidth]{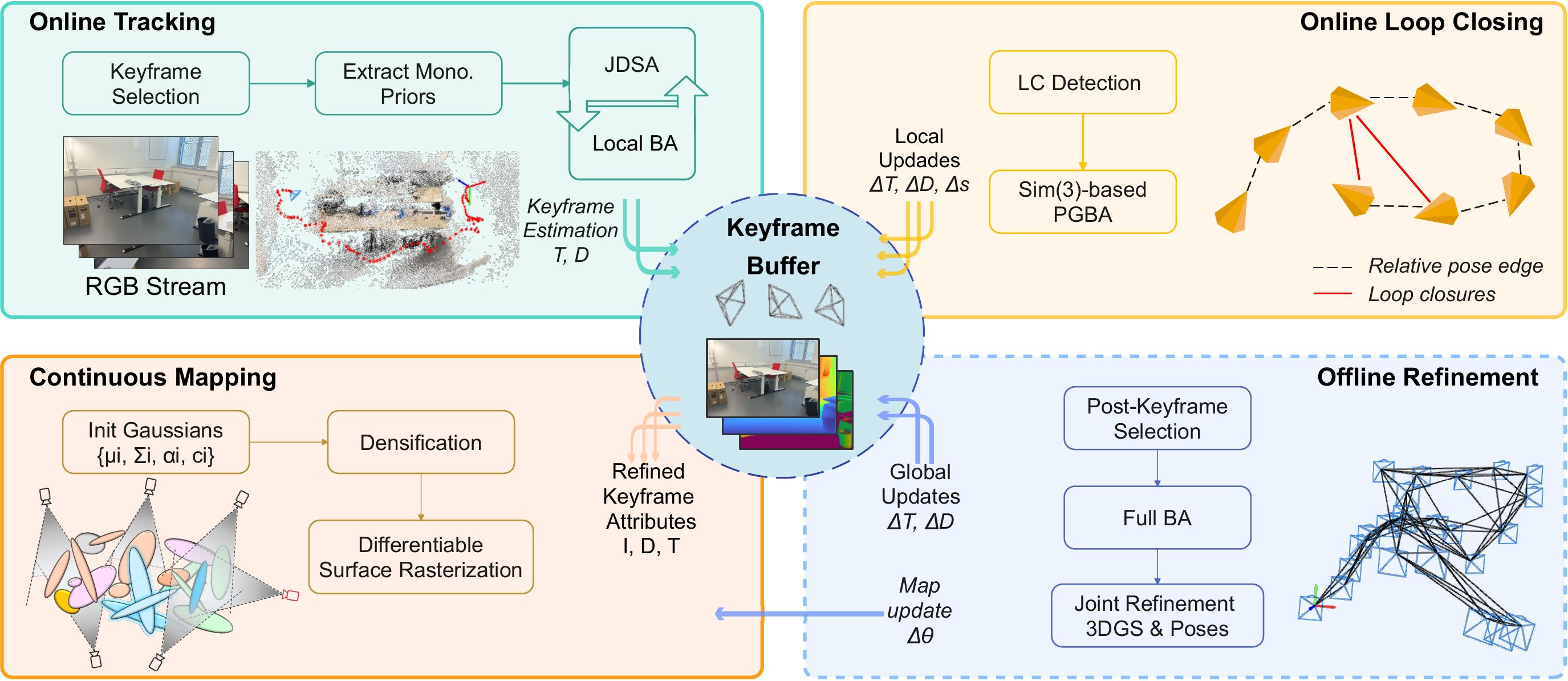}
\caption{\textbf{System Overview}: Our framework consists of four key stages: online camera tracking, online loop closing, online mapping, continuous mapping, and offline refinement. The camera tracking is performed using a recurrent-network-based approach to estimate camera poses $\mathbf{T}$ and generate depth maps $\mathbf{D}$ from RGB input. Depth priors are incorporated into the tracking process through our proposed Joint Depth and Scale Alignment (JDSA) strategy improving depth estimation accuracy. For 3D scene representation, we use 3DGS to model scene geometry, enabling efficient online map updates. These updates are integrated with $Sim(3)$-based pose graph Bundle Adjustment (BA) for online loop closing, allowing for scale drift correction via scale updates $\Delta \mathbf{s}$, and achieving both fast updates and high-quality rendering. In the offline refinement stage, camera poses and scene geometry undergo full BA, followed by joint optimization of Gaussian primitives and camera poses to further enhance global consistency.}
\label{fig:overview}
\vspace*{-3mm}
\end{figure*}

\section{Methods}
Our system is designed to enable fast and accurate camera tracking and scene reconstruction from monocular RGB input. As illustrated in Fig.~\ref{fig:overview}, the system comprises four key components: an online tracker, an online loop closing module, a continuous mapper, and an offline refinement stage. The online camera tracker (Sec.~\ref{sec:onlinetracking}) leverages a learning-based dense SLAM frontend to estimate camera poses and depth maps. Global consistency and real-time performance are achieved through the online loop closure module (Sec.~\ref{sec:loopclosing}), which combines loop closure detection with efficient Pose Graph Bundle Adjustment (PGBA). For scene representation, we employ 3DGS (Sec.~\ref{sec:3dscene}), enabling efficient online map construction, updates, and high-quality rendering. The offline refinement stage (Sec.~\ref{sec:offlinerefine}) enhances reconstruction quality through full BA and joint optimization of Gaussian map and camera poses ensures optimal global consistency. The final mesh is generated by fusing rendered depth maps through TSDF fusion.

\subsection{Comparison to HI-SLAM}
The current HI-SLAM2 system represents a significant advancement over our previous work HI-SLAM~\cite{zhang2024hi}, with improvements across multiple aspects that enhance tracking accuracy and reconstruction quality substantially. The key improvements can be summarized as follows:

\begin{itemize}
   \item \textbf{Depth Prior Integration:} We propose a novel spatially-adaptive scale-grid alignment strategy that effectively addresses nonlinear scale distortions in monocular depth priors. In contrast to HI-SLAM's single scale alignment, our 2D grid-based method with bilinear interpolation accommodates spatially varying distortions. This is achieved without introducing dependencies between depth pixels, as each depth value is treated as an independent variable. This design preserves the efficiency of solving the optimization problem using the Schur complement. The improved scale alignment enhances the accuracy of depth estimation, thereby facilitating Gaussian initialization and optimizing the map reconstruction with depth supervision.

   \item \textbf{Map Representation:} We replace HI-SLAM's neural implicit field representation with 3DGS, transitioning from an implicit to explicit representation. This change provides several benefits: (1) significantly faster rendering, (2) efficient online map updates through direct Gaussian primitive updates rather than neural network weight optimization, (3) incremental map growth without predefined scene boundaries, and (4) enhanced geometry preservation as evidenced by the 1.54cm improvement in reconstruction accuracy on the Replica dataset (Table~\ref{tab:recon_replica}).
   
   \item \textbf{Hierarchical Optimization:} Our system employs a multi-stage optimization pipeline that includes online tracking with local BA, online loop closure with $Sim(3)$-based PGBA, and global full BA. Last but not least, the joint refinement of both Gaussian map parameters and camera poses, which couples the tracking frontend with the mapping backend more tightly. In contrast, HI-SLAM only performs pose optimization in the tracking frontend, which can lead to inconsistencies between estimated poses and map geometry. This hierarchical approach reduces the Absolute Trajectory Error (ATE) by 29.3\% compared to online tracking alone (Table~\ref{tab:ablation_ate}) on the Replica dataset, yielding a globally more consistent reconstruction.

\end{itemize}

\subsection{Online Tracking}\label{sec:onlinetracking}
Our online tracking module builds upon a learning-based dense visual SLAM method~\cite{teed2021droid} to estimate camera poses and depth maps of keyframes. By leveraging dense per-pixel information through a recurrent optical flow network, our system can robustly track the camera in challenging scenarios, such as low-textured environments and fast movements. To match per-pixel correspondences among all overlapping frames, we construct a keyframe graph $(\mathcal{V}, \mathcal{E})$ which represents the co-visibility relationships between every pair of keyframes. The graph nodes $\mathcal{V}$ correspond to keyframes, each containing a pose $\mathbf{T} \in SE(3)$ and an estimated depth map $\mathbf{d}$. Graph edges $\mathcal{E}$ connect keyframes with sufficient overlap, determined by their optical flow correspondences. To synchronize the estimated states with other modules aiding continuous mapping and online loop closing, a keyframe buffer is maintained to store the information of all keyframes and their respective states. 

The tracker begins with keyframe selection where each incoming frame is assessed to determine if it should be selected as a keyframe. This decision is based on the average flow distance relative to the last keyframe calculated through a single pass of the optical flow network~\cite{teed2020raft} and a predefined threshold $d_{flow}$. For selected keyframe, we extract the monocular priors, including depth and normal priors, through a pretrained neural network~\cite{eftekhar2021omnidata}. While the depth priors are used directly by the tracker module to facilitate depth estimation, the normal priors are used by the scene representation mapper for 3D Gaussian map optimization as extra geometry cues. 

Following~\cite{teed2021droid}, we initialize the system state after collecting $N_{init}=12$ keyframes. The initialization performs bundle adjustment (BA) on a keyframe graph, where edges connect keyframes within an index distance of 3, ensuring sufficient overlap for reliable convergence. Since a monocular system does not have an absolute scale, we normalize the scale by setting the mean of all keyframe depths to one. This scale is then hold as the system scale by fixing the poses of the first two keyframes in subsequent BA optimizations. Afterwards, each time a new keyframe is added, we perform local BA to estimate the camera poses and depth maps of the keyframes in the current keyframe graph. Edges between the new keyframe and neighboring keyframes with sufficient overlap are added to the graph. With the optical flow prediction $\mathbf{f}$, the reprojection error is minimized by using the flow-predicted target, denoted as $\mathbf{\check{p}}_{ij} = \mathbf{p}_i + \mathbf{f}$, and the current reprojection induced by camera poses and depths as source. The local BA optimization problem can be formulated as:
\begin{equation}\label{equ:1}
\argminA_{\mathbf{T},\mathbf{d}} \sum_{(i,j) \in \mathcal{E}}^{} \| \mathbf{\check{p}}_{ij} - \Pi(\mathbf{T}_{ij} \Pi^{-1}(\mathbf{p}_i, \mathbf{d}_i)) \|_{\Sigma_{ij}}^2
\end{equation}
where $\mathbf{T}_{ij}=\mathbf{T}_{j} \cdot \mathbf{T}_{i}^{-1}$ denotes the rigid body transformation from keyframe $i$ to keyframe $j$, and $\mathbf{d}_i$ refers to the depth map of keyframe $i$ in inverse depth parametrization, $\Pi$ and $\Pi^{-1}$ represent the camera projection and back-projection functions, respectively. $\Sigma_{ij}$ is a weight matrix with diagonals representing the prediction confidences from the optical flow network. The confidence effectively ensures the robustness of the optimization by reducing the influence of the outliers caused by occlusions or low-texture regions. Depth estimates in under-confident regions, where the depth cannot be accurately estimated, are further refined using monocular depth priors in the subsequent step.

\textbf{Incorporate Monocular Depth Prior}: To overcome the challenge of depth estimation in difficult areas such as low-textured or occluded regions, we incorporate the easy-to-obtain monocular depth priors~\cite{eftekhar2021omnidata} into the online tracking process. In the RGB-D mode of~\cite{teed2021droid}, depth observations are directly used to compute the mean squared error during BA optimization. However, we can not directly follow the same manner because predicted monocular depth priors exhibit inconsistent scales. To address this, \cite{zhang2024hi} proposes estimating a depth scale and an offset for each depth prior as optimization parameters. Although this approach helps align an overall prior scale, we found that it is not sufficient to fully correct the scale distortions inherent in monocular depth priors. 

\begin{figure}[t!]
\centering
\includegraphics[width=0.496\textwidth]{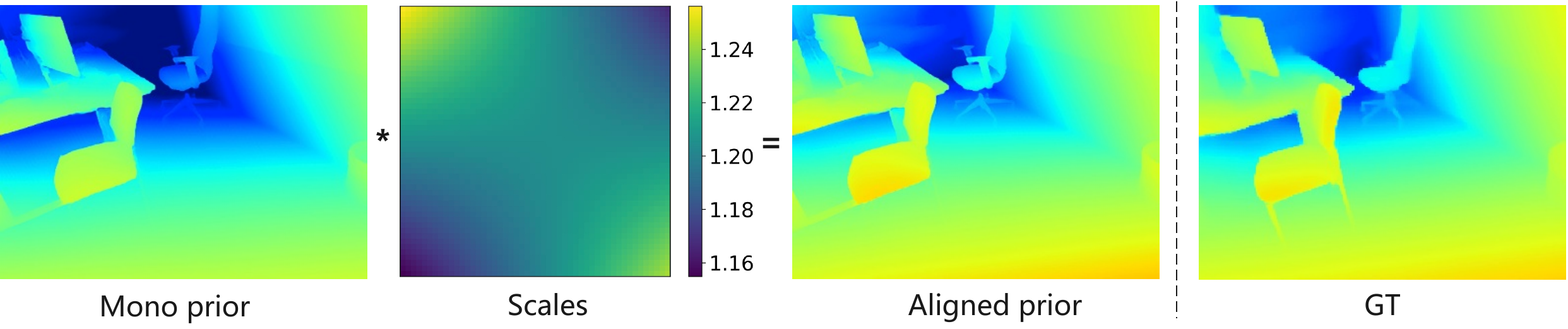}
\vspace*{-5mm}
\caption{Example of scale alignment of monocular depth. }
\label{fig:jdsa}
\end{figure}

To further improve this, we propose estimating a 2D depth scale grid with coefficients $\mathbf{s}_i$ of dimension $(m,n)$ for each depth prior $\check{\mathbf{d}}_i$. The depth scale at every pixel can be obtained by bilinear interpolation $Bi(\mathbf{p}_i, \mathbf{s}_i)$ on the grid based on its four surrounding grid coefficients. This spatially-varying scale formulation makes it more flexible to align the prior depth with the estimated depth by BA and helps to reduce the influence of noise in the depth prior. Using the sampled depth scales, the scale-aligned depth prior can be obtained as $\check{\mathbf{d}}_i \cdot Bi(\mathbf{p}_i, \mathbf{s}_i)$. Then we formulate the depth prior factor $r_{d}$ as follows:
\begin{equation}\label{equ:dprior}
	r_{d} = \|\check{\mathbf{d}}_i \cdot Bi(\mathbf{p}_i, \mathbf{s}_i) - \mathbf{d}_i \|^2 
\end{equation}
The grid resolution is set to $2 \times 2$. Higher resolutions may introduce instability, particularly in low-texture regions where optical flow predictions have low confidence, resulting in insufficient weighting in the reprojection error term for each grid tile. The scale coefficients are initially set to ones. After the system converges, the scale coefficients for new depth priors are initialized using those from the previous depth priors.

As reported in~\cite{zhang2024hi}, directly incorporating the depth prior factor into the BA optimization, i.e. jointly optimizing the camera poses, depths, and scale coefficients, can make the system prone to scale drift and hinder convergence. To address this, similar to the approach in~\cite{zhang2024hi}, we introduce a joint depth and scale alignment (JDSA) module to estimate the prior scales separately with the following objective:
\begin{equation}\label{equ:JDSA}
   \begin{aligned}
  \argminA_{\mathbf{s}, \mathbf{d}} &\sum_{(i,j) \in \mathcal{E}}^{} \| \check{\mathbf{p}}_{ij} - \Pi(\mathbf{T}_{ij} \Pi^{-1}(\mathbf{p}_i, \mathbf{d}_i)) \|_{\Sigma_{ij}}^2 + \\
    &\ \ \sum_{i \in \mathcal{V}}^{} \ \ \|\check{\mathbf{d}}_i \cdot Bi(\mathbf{p}_i, \mathbf{s}_i) - \mathbf{d}_i \|^2 
   \end{aligned}
\end{equation}
By interleaving the JDSA optimization with the local BA optimization, we ensure that the system scale remains stable and the depth prior is well-aligned, providing depth estimation with a better initial guess. We use the damped Gauss-Newton algorithm to solve the optimization problem. For the sake of the optimization efficiently, we separate scale and depth variables as follows:
\begin{equation}
   \begin{pmatrix} \mathbf{B} & \mathbf{E} \\ \mathbf{E}^T & \mathbf{C} \end{pmatrix} \begin{pmatrix} \Delta \mathbf{s} \\ \Delta \mathbf d \end{pmatrix} = \begin{pmatrix} \mathbf v \\ \mathbf w \end{pmatrix} \qquad
\end{equation}
where $\mathbf{B}$, $\mathbf{E}$, $\mathbf{C}$ are the blocks of the Hessian matrix and $\mathbf{v}$, and $\mathbf{w}$ are the gradient vector of the linearized system. Since the dimension of matrix $\mathbf{B}$ is much smaller than $\mathbf{C}$, we can solve the system efficiently by first solving for $\Delta \mathbf{s}$ and then $\Delta \mathbf{d}$ using the Schur complement.
\begin{equation}
   \begin{aligned}
      \Delta \mathbf{s} &= (\mathbf B - \mathbf E \mathbf C^{-1} \mathbf E^T)^{-1}(\mathbf v - \mathbf E \mathbf C^{-1} \mathbf w) \\
      \Delta \mathbf{d} &= \mathbf C^{-1}(\mathbf w - \mathbf E^T\Delta \mathbf{s})
   \end{aligned}
\end{equation}
Matrix $\mathbf{C}$ is diagonal since the scale alignment in Eq.~\ref{equ:dprior} is applied to the depth prior rather than the depth variables. This preserves the independence between the depth variables allowing us to invert $\mathbf{C}$ efficiently as $\mathbf{C}^{-1}=diag(\frac{1}{c_1},..,\frac{1}{c_n})$.
Fig.~\ref{fig:jdsa} shows an example of the scale alignment for monocular depth priors. Note that the estimated spatially-varying scales result in well-aligned depth prior with respect to the ground truth depth.

\subsection{Online Loop Closing}\label{sec:loopclosing}
While our online tracker can robustly estimate camera poses, measurement noise inevitably accumulates over time and travel distance, which leads to pose drift. Additionally, monocular systems are prone to scale drift due to inherent scale unobservability. To correct both pose and scale drifts and enhance the global consistency of the 3D map, our online loop closing module searches for potential loop closures and performs global optimization on the entire history of keyframes using a $Sim(3)$-based PGBA first proposed in~\cite{zhang2024hi}. 

\textbf{Loop Closure Detection}: Loop closure detection is performed in parallel to the online tracking. For each selected new keyframe, we calculate the optical flow distances $d_{of}$ between the new keyframe and all previous keyframes. We define three criteria to select loop closure candidates. First, $d_{of}$ must fall below a predefined threshold $\tau_{flow}$, ensuring sufficient co-visibility for reliable convergence in recurrent flow updates. Second, orientation differences based on current pose estimations should remain within a threshold $\tau_{ori}$. Finally, the frame index difference must exceed a minimum threshold $\tau_{temp}$ beyond the current local BA window. When all criteria are met, we add edges connecting the keyframe pairs in forward and revert re-projection directions in our keyframe graph.

\textbf{$Sim(3)$-Based Pose Graph Bundle Adjustment}: When loop closure candidates are identified, inspired by the efficiency of PGBA in~\cite{wei2023bamf,zhang2024hi}, we choose pose graph BA over full BA to balance computational efficiency with accuracy. To address scale drift, we adopt $Sim(3)$ representations for keyframe poses, enabling per-keyframe scale correction as proposed in~\cite{strasdat2010scale}. Before each optimization run, we convert the latest pose estimates from $SE(3)$ to $Sim(3)$ group and initialize scales with ones. The pixel warping step follows Eq.~\ref{equ:1}, with the $SE(3)$ transformation replaced by a $Sim(3)$ transformation.

Constructing the pose graph involves connecting poses through relative pose edges. Following~\cite{wei2023bamf,zhang2024hi}, we derive relative poses from dense correspondences of inactive reprojection edges which are retained when their associated keyframes leave the sliding window of local BA. These dense correspondences offer a reliable basis for computing relative poses because they have been refined for multiple iterations when they are active in the sliding window. The reprojection error term from Eq.~\ref{equ:1} is used, but here the optimization focuses solely on relative poses $\mathbf{\check{T}}_{ij}$ under the assumption that depth estimates are accurate. To incorporate uncertainty, we estimate variances $\Sigma^{rel}_{ij}$ for the relative poses based on the adjustment theory~\cite{niemeier2008ausgleichungsrechnung} as:
\begin{equation}
   \Sigma^{rel}_{ij} = (\mathbf{J} \Delta \mathbf{T_{ij}} - \mathbf{r})^T \Sigma_{ij} (\mathbf{J} \Delta \mathbf{T_{ij}} - \mathbf{r}) (\mathbf{J}^T \Sigma_{ij} \mathbf{J})^{-1}
\end{equation}
where $\mathbf{J}$, $\mathbf{r}$, and $\Delta \mathbf{T_{ij}}$ are the Jacobian, reprojection residuals, and relative pose update from the previous iteration, respectively. These variances serve as weights in PGBA. The final objective of PGBA is to minimize the sum of relative pose factors and reprojection factors:
\begin{equation}\label{equ:pgba}
   \begin{aligned}
  \argminA_{\mathbf{T}, \mathbf{d}} &\sum_{(i,j) \in \mathcal{E}^*}^{} \| \mathbf{\check{p}}_{ij} - \Pi(\mathbf{T}_{ij} \Pi^{-1}(\mathbf{p}_i, \mathbf{d}_i)) \|_{\Sigma_{ij}}^2 + \\
    & \sum_{(i,j) \in \mathcal{E}^{+}}^{} \| \log(\mathbf{\check{T}}_{ij} \cdot \mathbf{T}_{i} \cdot \mathbf{T}_{j}^{-1}) \|_{\Sigma^{rel}_{ij}}^2
   \end{aligned}
\end{equation}
where $\mathcal{E}^*$ represents detected loop closures, and $\mathcal{E}^+$ denotes the set of relative pose factors. To ensure the convergence of the optimization and account for potential outliers in relative pose factors, we apply a damped version of Gauss-Newton algorithm to find the optimal solution as follows:
\begin{equation}
   \mathbf{H} = \mathbf{H} + \epsilon \cdot \mathbf{I} + \lambda \cdot \mathbf{H} 
\end{equation}
where $\mathbf{H}$ denotes the Hessian matrix. The damping factor $\epsilon=10^{-4}$ and regularization factor $\lambda=10^{-1}$ serve two critical functions: preventing convergence to local minima and improving numerical conditioning, while maintaining rapid convergence. Following optimization, we convert the optimized poses back to the $SE(3)$ for subsequent tracking. The depth maps are scaled according to the corresponding $Sim(3)$ pose transformations. Additionally, as detailed in Sec.~\ref{sec:3dscene}, we update the 3D Gaussian primitives based on the pose updates of their anchor keyframes.

\subsection{3D Scene Representation}\label{sec:3dscene}
We adopt 3DGS~\cite{kerbl20233d} as our scene representation modeling scene appearance and geometry. Unlike implicit neural representations such as NeRF, 3DGS provides an explicit representation that enables efficient online map updates and high-quality rendering. The scene is represented by a set of 3D anisotropic Gaussians $\mathcal{G}=\{g_i\}_{i=1}^{M}$, where each 3D Gaussian unit is defined as:
\begin{equation}
g_i(\mathbf{x}) = e^{-(\mathbf{x}-\mathbf{\mu}_i)^\top \mathbf{\Sigma}_i^{-1}(\mathbf{x}-\mathbf{\mu}_i)},
\end{equation}
where $\mathbf{\mu}_i \in \mathbb{R}^3$ denotes the Gaussian mean and $\mathbf{\Sigma}_i \in \mathbb{R}^{3 \times 3}$ represents the covariance matrix in world coordinates. The covariance matrix $\mathbf{\Sigma}_i$ is decomposed into orientation $\mathbf{R}_i$ and scale $\mathbf{S}_i=diag\{s_i\} \in \mathbb{R}^{3 \times 3}$, such that $\mathbf{\Sigma}_i = \mathbf{R}_i \mathbf{S}_i \mathbf{S}_i^T \mathbf{R}_i^T$. Each Gaussian also carries attributes for opacity $o_i \in [0,1]$ and color $\mathbf{c}_i \in \mathbb{R}^3$. Unlike the original 3DGS~\cite{kerbl20233d}, we simplify the color representation by using direct RGB values instead of spherical harmonics, reducing optimization complexity. To handle view-dependent color variations, we employ exposure compensation during the offline refinement stage (Sec.~\ref{sec:offlinerefine}).

The rendering process projects these 3D Gaussians onto the image plane using perspective transformation:
\begin{equation}
\begin{aligned}
& \mathbf{\mu}_i' = \pi(\mathbf{T}_i \cdot \mathbf{\mu}_i),  \\
& \mathbf{\Sigma}_i' = \mathbf{J} \mathbf{W} \mathbf{\Sigma}_i \mathbf{W}^T \mathbf{J}^T
\end{aligned}
\end{equation}
where $\mathbf{J}$ represents the Jacobian of the perspective transformation and $\mathbf{W}$ denotes the rotation matrix of keyframe pose $\mathbf{T}_i$. After depth-based sorting of the projected 2D Gaussians, pixel colors and depths are computed through $\alpha$-blending along each ray from near to far:
\begin{equation}
\begin{aligned}
   & \hat{C} = \sum_{i\in \mathcal{N}} c_i\alpha_i \prod_{j=1}^{i-1}(1-\alpha_j), \\ 
   & \hat{D} = \sum_{i\in \mathcal{N}} d_i\alpha_i \prod_{j=1}^{i-1}(1-\alpha_j)
\end{aligned}
\end{equation}
where $\mathcal{N}$ represents the set of Gaussians intersecting the ray, $c_i$ is the color of the $i$-th Gaussian, and $\alpha_i$ represents the pixel translucency calculated by evaluating the opacity of i-th Gaussian at the intersection point. 

\textbf{Unbiased Depth}: Previous works~\cite{keetha2024splatam, matsuki2024gaussian} directly use the depth at the Gaussian mean, which introduces estimation biases when rays intersect the Gaussian at points distant from its mean. Following~\cite{zhang2024rade}, we compute an unbiased depth by determining the actual ray-Gaussian intersection point along the ray direction. This depth is calculated by solving the planar equation at the intersection of the ray and Gaussian surface. Since all rays from the same viewpoint that intersect a given Gaussian are co-planar, the intersection equation needs to be solved only once per Gaussian. This approach maintains the computational efficiency of splat-based rasterization while significantly improving depth accuracy. We demonstrate the benefits of this unbiased depth computation through ablation studies in Sec.~\ref{sec:ablation}.


\textbf{Map Update}:The map update process adjusts the 3D Gaussian units based on the updates of keyframe pose to ensure global consistency of the 3D map. This update happens both online during the $Sim(3)$-based PGBA and offline during the global full BA. To enable rapid and flexible updates to the 3D scene representation, we deform the mean, orientation, and scale of each Gaussian unit. Specifically, means and orientations are transformed according to the relative $SE(3)$ pose change between the previous and updated keyframes, while scales are adjusted using the scale factors derived from the $Sim(3)$ pose representation.

The update equations for each Gaussian unit are:
\begin{equation}
\begin{aligned}
& \mathbf{\mu}_{j}' = (\mathbf{T}_{i}'^{-1} \cdot \mathbf{T}_{i} \cdot \mathbf{\mu}_{j}) / s_{i}, \\ 
& \mathbf{R}_{j}' = \mathbf{R}_{i}'^{-1} \cdot \mathbf{R}_{i} \cdot \mathbf{R}_{j}, \\
& s_{j}' = s_{i} \cdot s_{j}
\end{aligned}
\end{equation}
where $\mathbf{\mu}_{j}'$, $\mathbf{R}_{j}'$, and $s_{j}'$ represent the updated mean, orientation, and scale of the $j$-th Gaussian, respectively. This transformation ensures that the geometric relationships between Gaussians are preserved while accommodating the refined keyframe poses, maintaining the accuracy and completeness of the 3D reconstruction.

\textbf{Exposure Compensation}: Real-world captures exhibit varying exposures across different views due to illumination changes and view-dependent reflectance. These variations introduce color inconsistencies that can significantly impact reconstruction quality. Following~\cite{matsuki2024gaussian,kerbl2024hierarchical}, we address this challenge by optimizing per-keyframe exposure parameters through a 3×4 affine transformation matrix. For a rendered image $\hat{I}$, the exposure correction is formulated as:
\begin{equation}
   \hat{I}' = \mathbf{A} \cdot \hat{I} + \mathbf{b}
\end{equation}
where $\mathbf{A}$ denotes the 3×3 color transformation matrix and $\mathbf{b}$ represents the 3×1 bias vector. During the offline refinement stage, these exposure parameters are jointly optimized alongside camera poses and scene geometry, as detailed in Sec.~\ref{sec:offlinerefine}.

\textbf{Map Management}: To ensure that newly observed regions are well represented, we initialize a set of 3D Gaussian primitives when each new keyframe is created to populate the Gaussian map. The initialization process begins by unprojecting the estimated depth map of the keyframe into 3D space. Specifically, each pixel’s depth value is back-projected to generate a 3D point, which serves as the mean $\mu_i$ of a new Gaussian primitive. The orientation is initialized as the unit orientation. The initial scale $\mathbf{S}_i$ is set by finding the average distance of nearest 3 neighbors to adapt to the local point density, while opacity $o_i$ is initialized to 0.5 to allow the optimization itself to update. Color attributes $\mathbf{c}_i$ are assigned based on the corresponding pixel’s RGB value from the keyframe. 
To maintain map compactness and prevent redundancy, we apply random downsampling with a factor $\psi$. This downsampling ensures computational efficiency while preserving enough spatial coverage.
To control map growth, we implement a pruning strategy that removes Gaussians with low opacity to eliminate redundant or insignificant primitives. We reset the opacity values every 500 iterations and perform interleaved densification and pruning every 150 iterations to balance map size and quality. A detailed analysis of map size evolution is presented in Sec.~\ref{sec:runtimemapsize}.

\textbf{Optimization Losses}: The 3DGS representation is optimized using a combination of photometric, geometric, and regularization losses. The photometric loss $\mathcal{L}_{c}$ measures the L1 difference between the exposure-compensated rendered image $\hat{I}'$ and the observed image $I$. The depth loss $\mathcal{L}_{d}$ computes the L1 difference between the rendered depth $\hat{D}$ and the estimated depth $\bar{D}$ from the interleaved BA and JDSA optimization:
\begin{equation}
   \mathcal{L}_{c} = \sum_{k \in \mathcal{K}} | \hat{I}_k' - I_k |, \ \mathcal{L}_{d} = \sum_{k \in \mathcal{K}} | \mathbf{\hat{D}}_k - \bar{\mathbf{D}}_k |
\end{equation}
where $\mathcal{K}$ denotes the keyframes in the local window during online mapping or all keyframes during offline refinement. To enhance geometric supervision, we incorporate normal priors into the optimization. The estimated normals are derived from rendered depth maps using cross products of depth gradients along image plane axes. The normal loss $\mathcal{L}_{n}$ is defined as a cosine embedding loss:
\begin{equation}
   \mathcal{L}_{n} = \sum_{k \in \mathcal{K}} | 1 - \mathbf{\hat{N}}_k^T \cdot \mathbf{\bar{N}}_k |
\end{equation}
To prevent artifacts due to excessively slender Gaussians, we apply a regularization term to the scale of the 3D Gaussians:
\begin{equation}
   \mathcal{L}_{s} = \sum_{i \in \mathcal{G}} | \mathbf{s}_i - \bar{s}_i |
\end{equation}
where $\bar{s}_i$ denotes the mean scale of the i-th Gaussian, penalizing ellipsoid stretching. The final loss combines these terms with appropriate weights as follows:
\begin{equation}
   \mathcal{L} = \lambda_{c} \mathcal{L}_{c} + \lambda_{d} \mathcal{L}_{d} + \lambda_{n} \mathcal{L}_{n} + \lambda_{s} \mathcal{L}_{s}
\end{equation}
where $\lambda_{c}$, $\lambda_{d}$, $\lambda_{n}$, and $\lambda_{s}$ are the respective weights. We optimize Gaussian parameters using the Adam optimizer~\cite{kingma2014adam}, performing 10 iterations per new keyframe.

\subsection{Offline Refinement}\label{sec:offlinerefine}
Following the online processing, we implement three sequential offline refinement stages to enhance global consistency and map quality: post-keyframe insertion, full BA, and joint pose and map refinement.

\begin{figure}[t!]
\includegraphics[width=0.49\textwidth,left]{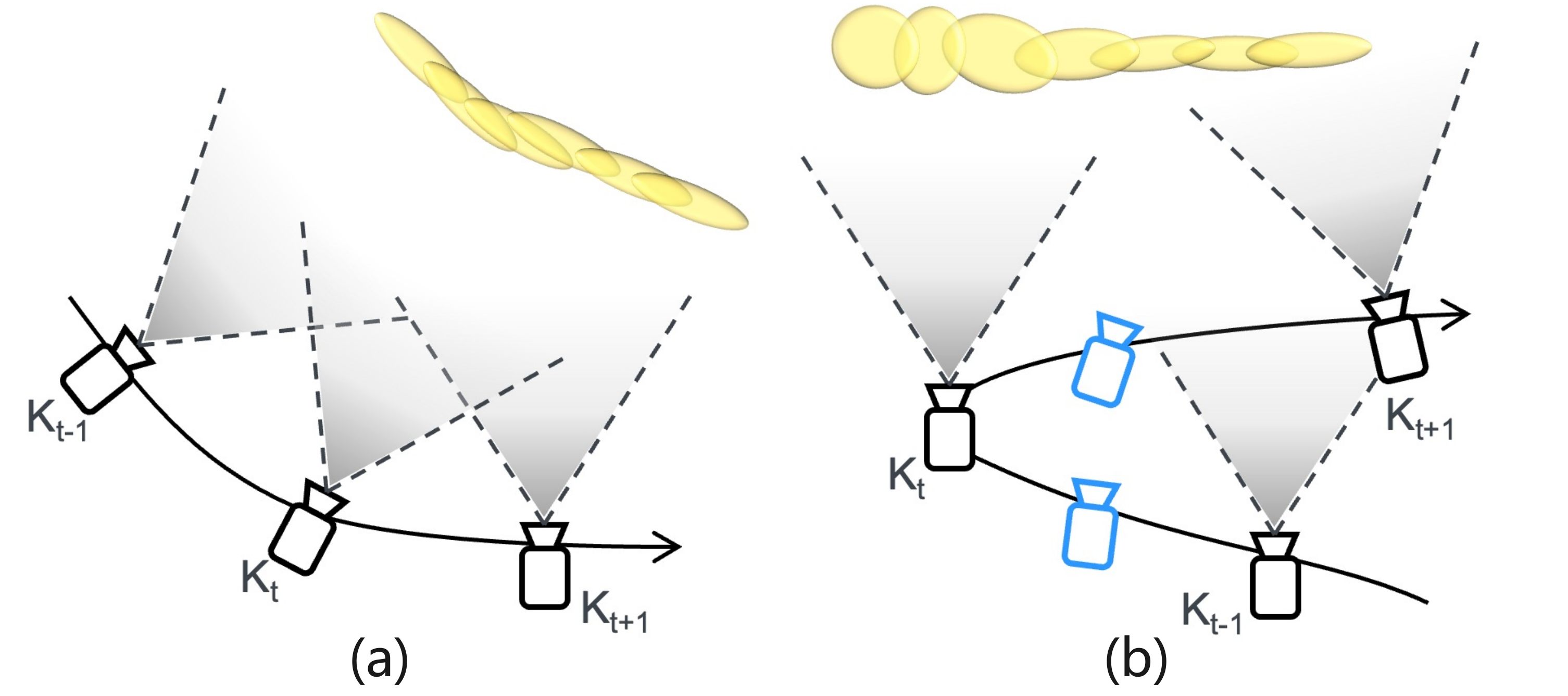}
\vspace*{-6mm}
\caption{View coverage analysis in two scenarios: (a) Optimal case where consecutive keyframes maintain sufficient overlap, ensuring proper multi-view coverage. (b) Suboptimal case where newly observed regions in keyframe $K_t$ lack adequate observations. Our system addresses this by inserting additional post-keyframes (shown in blue) to enhance view coverage.}
\label{fig:post-keyframe}
\end{figure}

\textbf{Post-Keyframe Insertion}: The first refinement stage identifies regions with insufficient view coverage, particularly areas near view frustum boundaries. These regions typically arise when forward camera motion is followed by backward rotational movement, as illustrated in Fig. \ref{fig:post-keyframe}. During online processing, keyframe selection relies on average optical flow between neighboring frames, as view coverage cannot be fully evaluated without future trajectory information. To identify under-observed regions in the offline stage, we project each keyframe's pixels onto its adjacent keyframes and quantify the percentage of pixels that fall outside the fields of view of neighboring keyframes. When this percentage exceeds a predetermined threshold, we flag the region as having insufficient observations. Additional keyframes are then inserted at these locations, and new Gaussian primitives are populated in the same manner as the keyframes inserted during the online process. This ensures more complete scene reconstruction and preserves critical details at scene boundaries.

\textbf{Full Bundle Adjustment}: While our online loop-closing module achieves global consistency through efficient $Sim(3)$-based PGBA, full BA further enhances system accuracy. PGBA offers superior computational efficiency compared to full BA, but introduces minor approximation errors when abstracting dense correspondences into relative pose edges. Specifically, PGBA computes reprojection factors only for loop closure edges, while full BA performs comprehensive optimization by re-computing reprojection factors in Eq.~\ref{equ:1} for all overlapping keyframe pairs, including both neighboring and loop closure frames. As demonstrated in Sec.~\ref{sec:ablation}, this improves the global consistency of camera poses and scene geometry at a finer granularity.

\textbf{Joint Pose and Map Refinement}: The final refinement stage jointly optimizes the Gaussian map and camera poses based on the results of full BA. While the online mapping stage limits optimization iterations per keyframe to maintain real-time performance, the offline refinement enables comprehensive optimization across all keyframes. To facilitate joint pose refinement, we compute pose Jacobians during rasterization-based rendering. Additionally, we also optimize per-keyframe exposure compensation parameters to ensure a better global color consistency. Unlike the full BA stage which employs the Gauss-Newton algorithm, this joint refinement step utilizes the Adam optimizer~\cite{kingma2014adam} with first-order gradient descent, leveraging our existing mapping pipeline.

\section{Experiments}
To evaluate the performance of the proposed system, we conducted extensive experiments on several challenging datasets, including the synthetic Replica dataset~\cite{straub2019replica} and the real-world datasets ScanNet~\cite{dai2017scannet}, Waymo~\cite{sun2020scalability}, ETH3D SLAM~\cite{Schops_2019_CVPR}, and ScanNet++~\cite{yeshwanth2023scannet++}. We begin by providing implementation details and evaluation metrics, followed by quantitative and qualitative comparisons on camera tracking accuracy, geometry, and appearance reconstruction quality against state-of-the-art baselines. Subsequently, we present ablation studies to analyze the impact of different design choices. Finally, we present the runtime performance and map size analysis.

\subsection{Implementation Details}
Our system is implemented using PyTorch~\cite{paszke2019pytorch} and CUDA for GPU acceleration, with evaluations performed on an Nvidia RTX 4090 GPU and Intel Core i9-12900K CPU. For optical flow and geometry prior prediction, we utilize pretrained models from~\cite{teed2021droid} and~\cite{eftekhar2021omnidata} respectively. For map refinement optimization, we use 2000 iterations for the Replica dataset and 26000 iterations for ScanNet and ScanNet++ datasets, ensuring fair comparison with existing methods. The loss weights of map optimization remain consistent across all experiments: color loss ($\lambda_{c}$) at 0.95, depth loss ($\lambda_{d}$) at 0.25, and scale loss ($\lambda_{s}$) at 10. The normal loss weight ($\lambda_{n}$) is set to 0.1 for the Replica dataset and increased to 0.5 for ScanNet and ScanNet++ datasets to enhance geometric supervision on real-world data. The downsampling factor $\psi$ is set to 32 across all experiments, providing an optimal balance between map size and quality.

\subsection{Datasets}
Replica Dataset~\cite{straub2019replica} provides synthetic indoor scenes with high-quality reconstructions, featuring complex geometry and textures. We evaluate using eight RGB-D sequences from~\cite{sucar2021imap}. The sequences have perfect camera poses and reconstructions make it ideal for benchmarking dense visual SLAM methods.
ScanNet Dataset~\cite{dai2017scannet} offers real-world RGB-D captures for 3D scene reconstruction. Following~\cite{zhang2024hi}, we use eight sequences for tracking evaluation and six additional sequences for geometry reconstruction assessment, using the RGB-D reconstructions as ground truth.
ETH3D SLAM~\cite{Schops_2019_CVPR} provides a diverse set of real-world RGB-D sequences with motion-capture ground truth poses, featuring challenging conditions such as extreme lighting variations and complete darkness.
Waymo Open dataset~\cite{sun2020scalability} provides real-world outdoor data with ground truth vehicle poses. The front-view camera images are used as input for our system. Following the evaluation protocol of~\cite{yu2025rgb}, we evaluate the tracking accuracy and rendering quality using 9 sequences.
ScanNet++~\cite{yeshwanth2023scannet++} presents a large-scale indoor dataset with laser-scanned ground truth, enabling evaluation of dense SLAM reconstruction quality. While the dataset includes both DSLR and iPhone captures, we specifically evaluate on the iPhone sequences, which present additional challenges due to their lower image quality.

\subsection{Evaluation Metrics}
We evaluate our system's performance across three key aspects. Camera tracking accuracy is quantified using Absolute Trajectory Error (ATE), measuring the precision of estimated camera poses. For geometry reconstruction quality, we adopt three metrics from~\cite{sucar2021imap}: average accuracy [cm], average completeness [cm], and completeness ratio [\%] (representing the percentage of reconstruction within 5cm of ground truth). For appearance quality assessment, we evaluate keyframe renderings using standard photometric metrics: PSNR (Peak Signal-to-Noise Ratio), SSIM (Structural Similarity Index), and LPIPS (Learned Perceptual Image Patch Similarity). In all result tables, we highlight performance rankings using: \colorbox{tabfirst}{first}, \colorbox{tabsecond}{second}, and \colorbox{tabthird}{third}.

\begin{figure}[t!]
\includegraphics[width=0.49\textwidth,left]{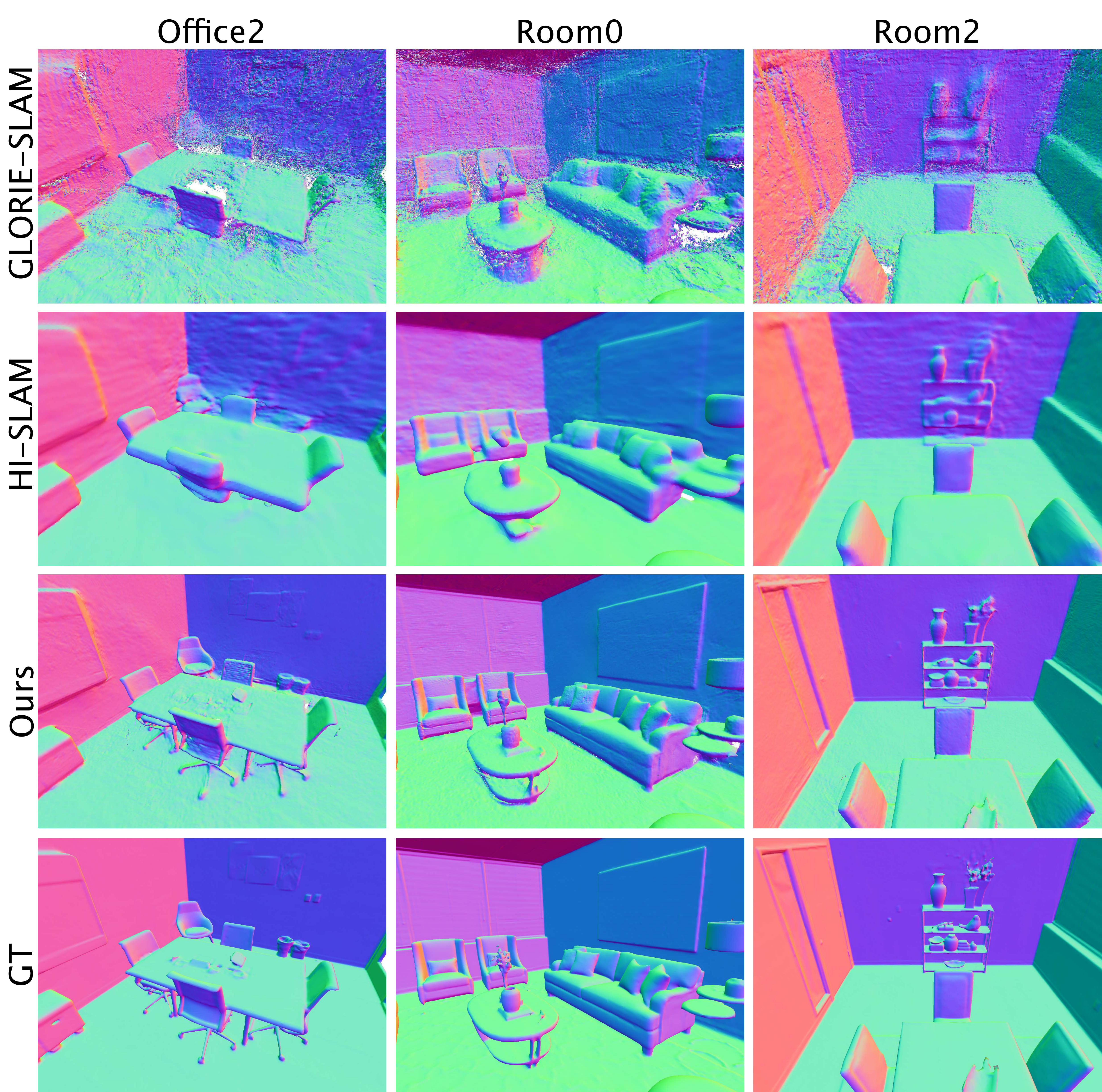}
\vspace*{-5mm}
\caption{Qualitative comparison on geometry reconstruction on Replica dataset.}
\label{fig:replica-quality}
\end{figure}

\subsection{Camera Tracking Accuracy}
\begin{table}[h]
\centering
\caption{Comparison of camera tracking accuracy for RGB and RGB-D methods on Replica dataset with results in [cm].}\label{tab:tracking_replica}
\setlength{\tabcolsep}{1.3pt}          %
\resizebox{0.494\textwidth}{!}{
\begin{tabular}{lcccccccccc}
\toprule
& Method & ro-0 & ro-1 & ro-2 & of-0 & of-1 & of-2 & of-3 & of-4 & Avg.\\
\midrule
\multirow{5}{*}{\rotatebox[origin=c]{90}{RGB-D input}}   &   NICE-SLAM\cite{zhu2022nice}   & 0.97 & 1.31 & 1.07 & 0.88 & 1.00 & 1.06 & 1.10 & 1.13 & 1.07 \\
    &   ESLAM\cite{johari2023eslam}   & 0.63 & 0.71 & 0.70 & 0.52 & 0.57 & 0.55 & 0.58 & 0.72 & 0.63 \\
    &   Point-SLAM\cite{sandstrom2023point}   & 0.61 & 0.41 & 0.37 & 0.38 & 0.48 & 0.54 & 0.69 & 0.72 & 0.52 \\
    &   SplatTAM\cite{keetha2024splatam}    & \rd0.31 & 0.40 & 0.29 & 0.47 & \,\fr0.27\, & 0.29 & \nd0.32 & 0.55 & 0.36 \\
    &   MonoGS\cite{matsuki2024gaussian}    & 0.44 & \rd0.32 & 0.31 & 0.44 & 0.52 & \,\fr0.23\, & \,\fr0.17\, & 2.25 & 0.58 \\
\midrule
\multirow{6}{*}{\rotatebox[origin=c]{90}{RGB input}}   &   DROID-SLAM\cite{teed2021droid}    & 0.34 & \,\fr0.13\, & 0.27 & \nd0.25 & 0.42 & 0.32 & 0.52 & \rd0.40 & \rd0.33 \\
    &   NICER-SLAM\cite{zhu2023nicer}   & 1.36 & 1.60 & 1.14 & 2.12 & 3.23 & 2.12 & 1.42 & 2.01 & 1.88 \\
    &   GLORIE-SLAM\cite{zhang2024glorie}    & \rd0.31 & 0.37 & \nd0.20 & \rd0.29 & \rd0.28 & 0.45 & 0.45 & 0.44 & 0.35 \\
    &   Splat-SLAM\cite{keetha2024splatam}    & \nd0.29 & 0.33 & \rd0.25 & \rd0.29 & 0.35 & 0.34 & 0.42 & 0.43 & 0.34 \\
    &   MGS-SLAM\cite{zhu2024mgs}     & 0.36 & 0.35 & 0.32 & 0.35 & \rd0.28 & \rd0.26 & \nd0.32 & \nd0.34 & \nd0.32 \\
    &   Ours    & \,\fr0.23\, & \nd0.22 & \,\fr0.19\, & \,\fr0.23\, & \,\fr0.27\, & \nd0.25 & 0.37 & \,\fr0.33\, & \,\fr0.26\, \\
\bottomrule
\end{tabular}}
\end{table}

\begin{table}[h]
\centering
\caption{Camera tracking accuracy for RGB and RGB-D methods on ScanNet dataset with results in [cm].} \label{tab:tracking_scannet}
\setlength{\tabcolsep}{1.5pt}          %
\resizebox{0.494\textwidth}{!}{
\begin{tabular}{lcccccccccc}
\toprule
& Method & 0000 & 0054 & 0059 & 0106 & 0169 & 0181 & 0207 & 0233 & Avg.\\
\midrule
\multirow{5}{*}{\rotatebox[origin=c]{90}{RGB-D input}}  &  NICE-SLAM\cite{zhu2022nice}   & 12.00 & 20.90 & 14.00 & 7.90 & 10.90 & 13.40 & \nd6.20 & 9.00 & 11.8 \\
&  ESLAM\cite{johari2023eslam}   & 7.30 & 36.30 & 8.50 & 7.50 & \nd6.50 & 9.00 & \,\fr5.70\, & \,\fr4.30\, & 10.6 \\
&  Co-SLAM\cite{wang2023co}   & 7.10 & 12.80 & 11.10 & 9.40 & \,\fr5.90\, & 11.80 & 7.10 & 6.10 & 8.90 \\
&  Point-SLAM\cite{sandstrom2023point}   & 10.20 & 28.00 & 7.80 & 8.70 & 22.20 & 14.80 & 9.50 & 6.10 & 14.3 \\
&  LoopSplat\cite{zhu2024loopsplat}   & \,\fr4.20\, & \,\fr7.50\, & \rd7.50 & 8.30 & \rd7.50 & 10.60 & 7.90 & 5.20 & 7.70 \\
\midrule
\multirow{5}{*}{\rotatebox[origin=c]{90}{RGB input}}  &  GO-SLAM\cite{zhang2023goslam}   & 5.90 & 13.30 & 8.30 & 8.10 & 8.40 & \rd8.30 & \rd6.90 & 5.30 & 8.10 \\
&  GLORIE-SLAM\cite{zhang2024glorie}    & \nd5.50 & \rd9.40 & 9.10 & \rd7.00 & 8.20 & \rd8.30 & 7.50 & \rd5.10 & \rd7.50 \\
&  Splat-SLAM\cite{keetha2024splatam}    & \rd5.57 & 9.50 & 9.11 & 7.09 & 8.26 & 8.39 & 7.53 & 5.17 & 7.58 \\
&  HI-SLAM\cite{zhang2024hi}    & 6.43 & 9.97 & \,\fr7.22\, & \,\fr6.56\, & 8.53 & \nd7.65 & 8.43 & 5.23 & \nd7.47 \\
&  Ours    & 5.82 & \nd8.64 & \nd7.30 & \nd6.80 & 8.25 & \,\fr7.41\, & 7.40 & \nd4.93 & \,\fr7.07\, \\
\bottomrule
\end{tabular}}
\end{table}
\begin{table}[ht!]
\centering
\caption{Camera tracking and rendering results on Waymo open dataset averaged over 9 sequences.} \label{tab:waymo}
\setlength{\tabcolsep}{2.0pt}          %
\resizebox{0.48\textwidth}{!}{
\begin{tabular}{rcccccc}
\toprule
Metrics & \makecell[c]{NICER-\\SLAM\cite{zhu2023nicer}} & \makecell[c]{GLORIE-\\SLAM\cite{zhang2024glorie}} & \makecell[c]{Photo-\\SLAM\cite{huang2024photo}} & \makecell[c]{MonoGS\cite{matsuki2024gaussian}} & \makecell[c]{OpenGS-\\SLAM\cite{yu2025rgb}} & \makecell[c]{\, \, Ours \, \,}\\
\midrule
ATE [m] $\downarrow$ & 19.59 & \nd0.536 & 19.95 & 8.529 & \rd0.839 & \fr0.457 \\
PSNR $\uparrow$ & 12.22 & 18.83 & 17.73 & \rd21.80 & \nd23.99 & \fr28.99 \\
SSIM $\uparrow$ & 0.622 & 0.702 & 0.741 & \rd0.780 & \nd0.800 & \fr0.872 \\
LPIPS $\downarrow$ & 0.726 & 0.572 & 0.674 & \rd0.577 & \nd0.434 & \fr0.219 \\
\bottomrule
\end{tabular}}
\end{table}

\begin{figure}[t!]
\includegraphics[width=0.48\textwidth, center]{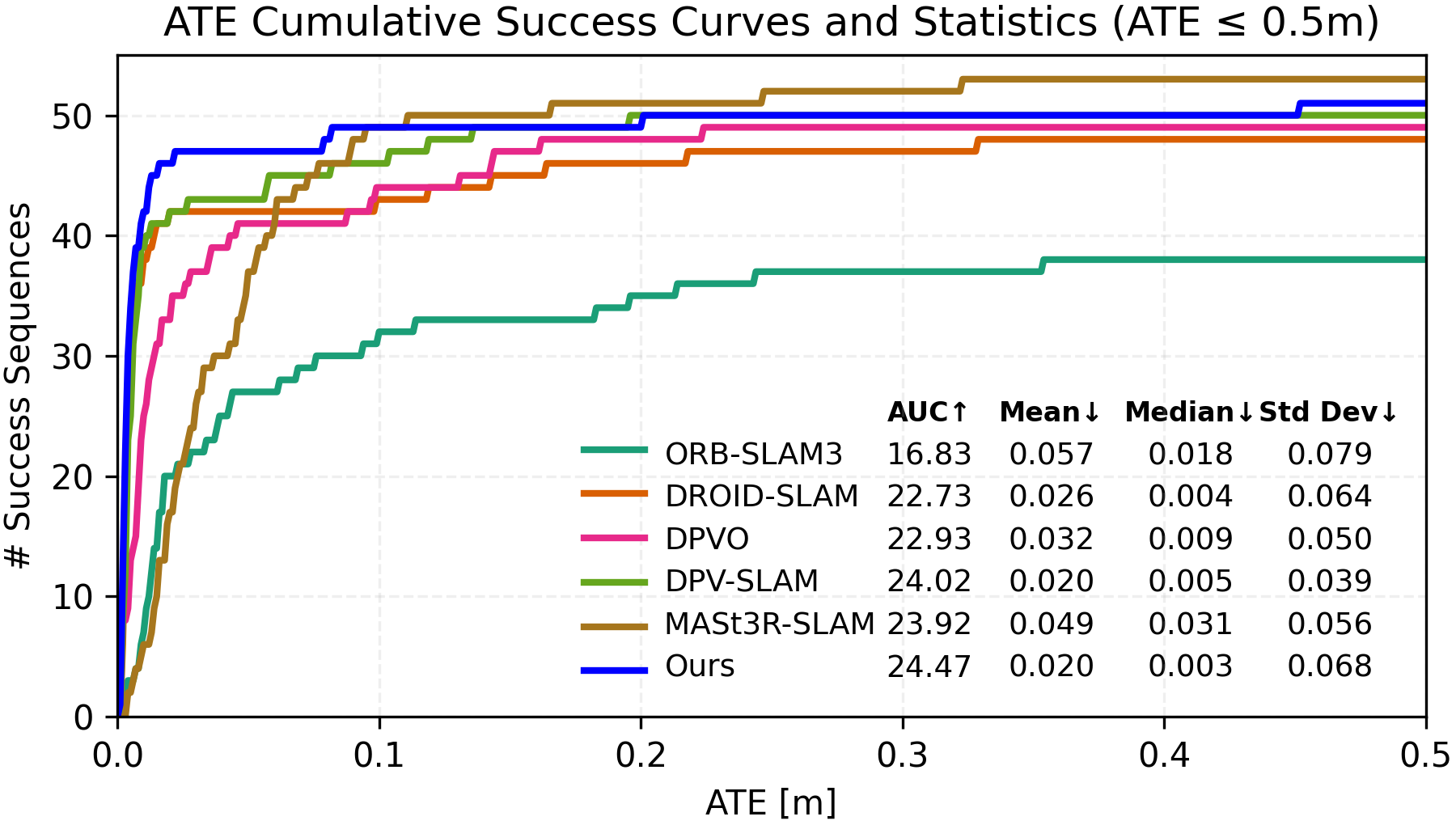}
\ifrevision
\else
\caption{Analysis of successful sequences relative to ATE error thresholds on the ETH3D SLAM dataset.}\label{fig:auc-plot}
\fi
\end{figure}
We evaluate the camera tracking accuracy of our system against state-of-the-art dense visual SLAM methods on indoor Replica, ScanNet, ETH3D SLAM datasets and outdoor Waymo datasets, including comparisons with RGB-D based approaches. The ATE results in Tables~\ref{tab:tracking_replica} and~\ref{tab:tracking_scannet} demonstrate the superior tracking accuracy of our system. 
RGB-D methods SplatTAM~\cite{keetha2024splatam} and MonoGS~\cite{matsuki2024gaussian}, despite having access to depth measurements for map-based tracking, achieve lower accuracy than hybrid approaches. Our system, along with Splat-SLAM~\cite{sandstrom2024splat}, represents the class of hybrid methods that effectively combine dense SLAM with deep learning foundations. The global BA of DROID-SLAM~\cite{teed2021droid} was enabled in all experiments. While it employs global BA, our additional global pose and map joint refinement further improves tracking accuracy beyond the baseline method.
We further compare our system on the ETH3D SLAM dataset with state-of-the-art sparse and dense methods, including ORB-SLAM3~\cite{campos2021orb}, DPVO~\cite{teed2023deep}, DPV-SLAM~\cite{lipson2024deep}, and MASt3R-SLAM~\cite{murai2024mast3r}. Figure~\ref{fig:auc-plot} illustrates the cumulative success curves based on ATE thresholds\revadd{ and ATE statistics}. As none of the methods can successfully track all sequences, we use the Area Under the Curve (AUC) metric to assess both accuracy and robustness with an upper ATE threshold of 0.5m. Our system achieves the highest AUC among all methods\revadd{ and lowest mean and median ATE}. Out of total 61 sequences, 6 sequences failed due to complete darkness, while 4 sequences encountered tracking failures caused by lighting changes and view occlusions. These limitations could potentially be addressed in the future by integrating place recognition for relocalization\revadd{ to reduce the standard deviation of ATE}.
Our evaluation on the Waymo Open dataset (Table~\ref{tab:waymo}) further validates our approach, where we achieve the lowest ATE among all competing methods. This highlights the ability of our system to generalize effectively to challenging large-scale outdoor environments with complex scene geometries that typically pose substantial difficulties for monocular systems.

\subsection{Geometry Reconstruction Quality}
\begin{table}[h]
\centering
\caption{Reconstruction evaluation on Replica dataset for implicit and explicit rgb methods. Ours surpass other methods especially large margin in accuracy.}.  \label{tab:recon_replica}
\setlength{\tabcolsep}{0.9pt}          %
\resizebox{0.493\textwidth}{!}{
\begin{tabular}{lclccccccccc}
\toprule
& Method & Metric & ro-0 & ro-1 & ro-2 & of-0 & of-1 & of-2 & of-3 & of-4 & Avg.\\
\midrule
\multirow{9}{*}{\rotatebox[origin=c]{90}{NeRF-based}}   &   \multirow{3}{*}{\makecell{NICER-\\SLAM\cite{zhu2023nicer}}}   &   \ml{1}{Acc.[cm] $\downarrow$}   & \rd2.53 & 3.93 & 3.40 & 5.49 & 3.45 & 4.02 & \rd3.34 & 3.03 & 3.65 \\
    &     &   \ml{1}{Comp.[cm] $\downarrow$}   & \fr3.04 & 4.10 & \nd3.42 & 6.09 & 4.42 & \fr4.29 & \rd4.03 & \nd3.87 & 4.16 \\
    &     &   Comp.Rat[\%]$\uparrow$   & \fr88.75 & 76.61 & \rd86.10 & 65.19 & 77.84 & 74.51 & \rd82.01 & \fr83.98 & 79.37 \\
\cmidrule(lr){2-12} 
    &   \multirow{3}{*}{\makecell{GO-\\\,SLAM\cite{zhang2023goslam}}}   &   \ml{1}{Acc.[cm] $\downarrow$}   & 4.60 & 3.31 & 3.97 & \rd3.05 & 2.74 & 4.61 & 4.32 & 3.91 & 3.81 \\
    &     &   \ml{1}{Comp.[cm] $\downarrow$}   & 5.56 & 3.48 & 6.90 & 3.31 & 3.46 & 5.16 & 5.40 & 5.01 & 4.79 \\
    &     &   Comp.Rat[\%]$\uparrow$   & 73.35 & 82.86 & 74.23 & 82.56 & \fr86.19 & 75.76 & 72.63 & 76.61 & 78.00 \\
\cmidrule(lr){2-12} 
    &   \multirow{3}{*}{\makecell{HI-\\SLAM\cite{zhang2024hi}}}   &   \ml{1}{Acc.[cm] $\downarrow$}   & 3.21 & 3.74 & 3.16 & 3.87 & 2.60 & 4.62 & 4.25 & 3.53 & 3.62 \\
    &     &   \ml{1}{Comp.[cm] $\downarrow$}   & \nd3.25 & \fr3.08 & 4.09 & 5.29 & 8.83 & 4.42 & 4.06 & \fr3.72 & 4.59 \\
    &     &   Comp.Rat[\%]$\uparrow$   & \rd86.99 & \fr87.19 & 80.82 & 72.55 & 72.44 & 80.90 & 81.04 & 82.88 & 80.60 \\
\midrule
\multirow{9}{*}{\rotatebox[origin=c]{90}{3DGS-based}}   &   \multirow{3}{*}{\makecell{GLORIE\\SLAM\cite{zhang2024glorie}}}   &   \ml{1}{Acc.[cm] $\downarrow$}   & 2.84 & \rd3.07 & \rd3.05 & \nd2.98 & \rd2.06 & \nd3.32 & \rd3.34 & \rd2.92 & \rd2.96 \\
    &     &   \ml{1}{Comp.[cm] $\downarrow$}   & 4.65 & 3.55 & \rd3.64 & \nd2.39 & \rd3.43 & 4.54 & 4.57 & 4.78 & \rd3.95 \\
    &     &   Comp.Rat[\%]$\uparrow$   & 81.96 & 85.78 & 84.50 & \nd88.82 & 85.07 & \fr82.09 & 80.41 & 81.04 & \rd83.72 \\
\cmidrule(lr){2-12} 
    &   \multirow{3}{*}{\makecell{Splat-\\SLAM\cite{sandstrom2024splat}}}   &   \ml{1}{Acc.[cm] $\downarrow$}   & \nd1.99 & \nd1.91 & \nd2.06 & 3.96 & \nd2.03 & \rd3.45 & \nd2.15 & \nd1.89 & \nd2.43 \\
    &     &   \ml{1}{Comp.[cm] $\downarrow$}   & 3.78 & \rd3.38 & \fr3.34 & \rd2.75 & \nd3.33 & \rd4.36 & \nd3.96 & \rd4.25 & \nd3.64 \\
    &     &   Comp.Rat[\%]$\uparrow$  & 85.47 & \nd86.88 & \nd86.12 & \rd87.32 & \rd85.17 & \rd81.37 & \nd82.25 & \rd82.95 & \nd84.69 \\
\cmidrule(lr){2-12} 
    &   \multirow{3}{*}{\makecell{Ours}}   &   \ml{1}{Acc.[cm] $\downarrow$}   & \fr1.35 & \fr1.40 & \fr1.87 & \fr1.40 & \fr1.18 & \fr1.94 & \fr1.70 & \fr1.70 & \fr1.57 \\
    &     &   \ml{1}{Comp.[cm] $\downarrow$}   & \rd3.33 & \nd3.27 & 3.66 & \fr2.07 & \fr3.23 & \fr4.29 & \fr3.84 & 4.26 & \fr3.49 \\
    &     &   Comp.Rat[\%]$\uparrow$   & \nd87.45 & \rd85.91 & \fr86.13 & \fr89.41 & \nd85.63 & \nd81.73 & \fr82.52 & \nd83.23 & \fr85.25 \\
\bottomrule
\end{tabular}}
\end{table}

\begin{figure*}[t!]
\captionsetup{justification=centering}
\includegraphics[width=0.98\textwidth,center]{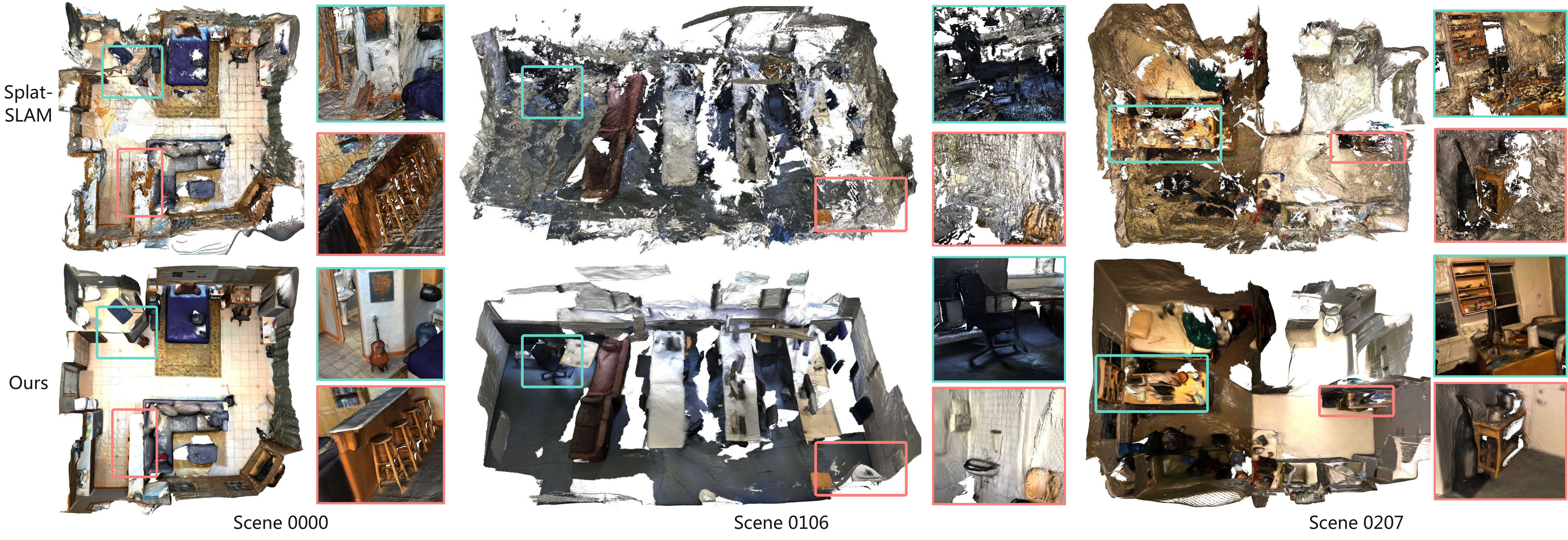}
\vspace*{-5mm}
\caption{Qualitative comparison on geometry and appearance reconstruction on ScanNet dataset.}
\label{fig:scannet-quality}
\end{figure*}

\begin{figure*}[t!]
\centering
\captionsetup{justification=centering}
\includegraphics[width=0.96\textwidth]{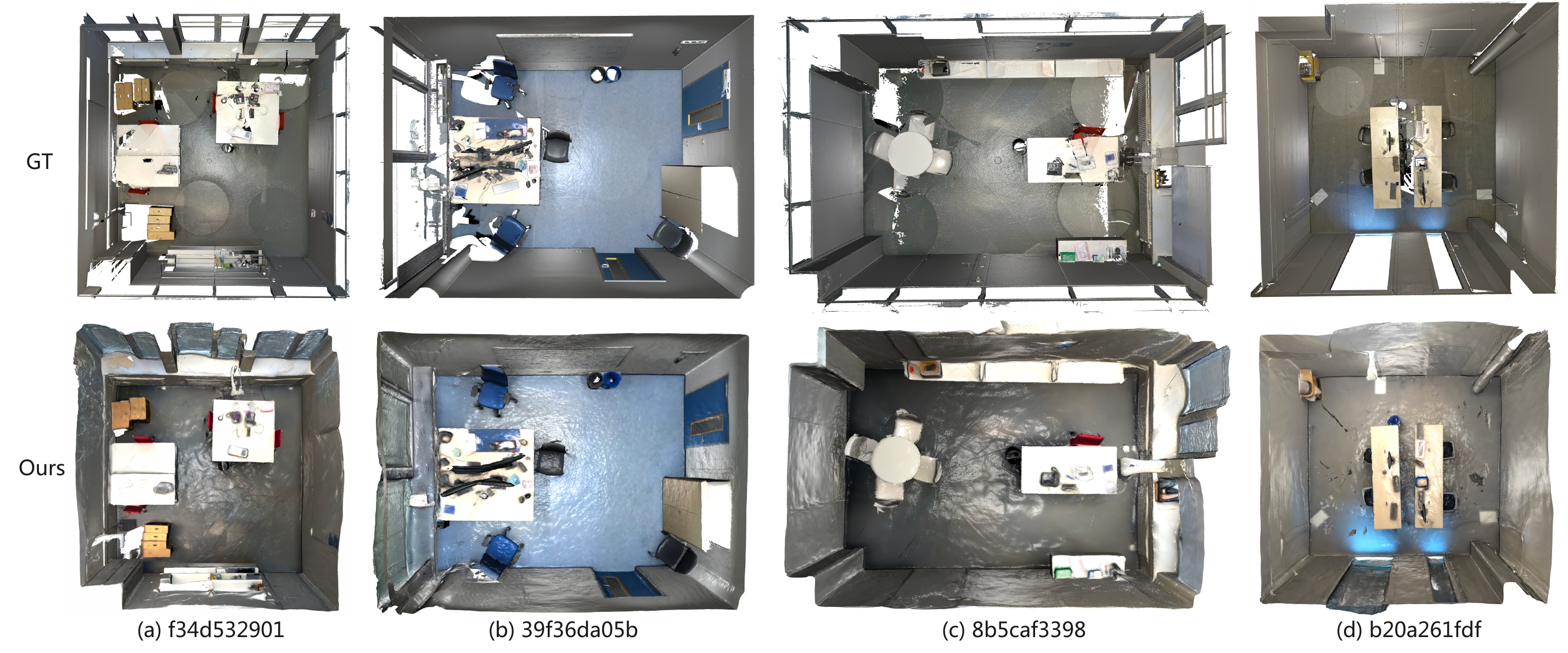}
\vspace*{-3mm}
\caption{Reconstructed meshes of four selected sequences on ScanNet++ Dataset.}
\label{fig:scannetpp}
\end{figure*}

In Table \ref{tab:recon_replica}, we evaluate our geometry reconstruction results against recent NeRF-based and 3DGS-based methods on the Replica dataset, demonstrating superior performance in both accuracy and completeness metrics. As illustrated in Fig.~\ref{fig:replica-quality}, our method produces smoother reconstructions while preserving fine geometric details compared to GLORIE-SLAM~\cite{zhang2024glorie} and HI-SLAM~\cite{zhang2024hi}. This is particularly evident in complex scene elements such as chair legs and shelf-mounted vases, where our results more closely match the ground truth. Qualitative comparisons on the ScanNet dataset (Fig.~\ref{fig:scannet-quality}) further highlight our advantages over Splat-SLAM~\cite{sandstrom2024splat}, showing more accurate geometry without floating artifacts and achieving better completeness. Additional qualitative results on the ScanNet++ dataset (Fig.~\ref{fig:scannetpp}) demonstrate our system's capability to fully reconstruct challenging scenes, including low-texture surfaces like floors and walls. Notably, our reconstructions even capture glass windows that are missing in the laser scanner ground truth.

\subsection{Appearance Reconstruction Quality}
\begin{table}[h]
\centering
\vspace{-3mm}
\caption{Rendering quality evaluations on Replica dataset for RGB and RGB-D methods.}\label{tab:replica_render}
\setlength{\tabcolsep}{1.0pt}          %
\resizebox{0.498\textwidth}{!}{
\begin{tabular}{lclccccccccc}
\toprule
& Method & Metric & ro-0 & ro-1 & ro-2 & of-0 & of-1 & of-2 & of-3 & of-4 & Avg.\\
\midrule
\multirow{10}{*}{\rotatebox[origin=c]{90}{RGB-D input}}    &    \multirow{3}{*}{\makecell{Point-\\SLAM\cite{sandstrom2023point}}}\,   &   \ml{1}{PSNR $\uparrow$}\,   & \,32.40\, & \,34.08\, & \,35.50\, & \,38.26\, & \,39.16\, & \,33.99\, & \,33.48\, & \,33.49\, & \,35.17\, \\
    &       &    \ml{1}{SSIM $\uparrow$}    & \nd0.97 & \fr0.98 & \fr0.98 & \fr0.98 & \fr0.99 & 0.96 & \rd0.96 & \fr0.98 & \fr0.98 \\
    &       &    LPIPS $\downarrow$    & 0.11 & 0.12 & 0.11 & 0.10 & 0.12 & 0.16 & 0.13 & 0.14 & 0.12 \\
\cmidrule(lr){2-12} 
    &    \multirow{3}{*}{\makecell{Splat\\TAM\cite{keetha2024splatam}}}    &    \ml{1}{PSNR $\uparrow$}    & \rd32.86 & 33.89 & 35.25 & 38.26 & 39.17 & 31.97 & 29.70 & 31.81 & 34.11 \\
    &       &    \ml{1}{SSIM $\uparrow$}    & \fr0.98 & \nd0.97 & \fr0.98 & \fr0.98 & \rd0.98 & \fr0.97 & 0.95 & 0.95 & \nd0.97 \\
    &       &    LPIPS $\downarrow$    & \nd0.07 & 0.10 & \rd0.08 & 0.09 & 0.09 & 0.10 & 0.12 & 0.15 & 0.10 \\
\cmidrule(lr){2-12} 
    &    \multirow{3}{*}{\makecell{Mono\\GS\cite{matsuki2024gaussian}}}    &    \ml{1}{PSNR $\uparrow$}    & \nd34.83 & \nd36.43 & \nd37.49 & \rd39.95 & \nd42.09 & \nd36.24 & \nd36.70 & \rd36.07 & \nd37.50 \\
    &       &    \ml{1}{SSIM $\uparrow$}    & 0.95 & 0.96 & \rd0.97 & 0.97 & \rd0.98 & 0.96 & \rd0.96 & 0.96 & 0.96 \\
    &       &    LPIPS $\downarrow$    & \nd0.07 & \nd0.08 & \rd0.08 & \rd0.07 & \rd0.06 & \rd0.08 & \rd0.07 & \rd0.10 & \rd0.07 \\
    \midrule
\multirow{10}{*}{\rotatebox[origin=c]{90}{RGB input}}    &    \multirow{3}{*}{\makecell{GLORIE\\-SLAM\cite{zhang2024glorie}}}    &    \ml{1}{PSNR $\uparrow$}    & 28.49 & 30.09 & 29.98 & 35.88 & 37.15 & 28.45 & 28.54 & 29.73 & 31.04 \\
    &       &    \ml{1}{SSIM $\uparrow$}    & \rd0.96 & \nd0.97 & 0.96 & \fr0.98 & \fr0.99 & \fr0.97 & \fr0.97 & \rd0.97 & \nd0.97 \\
    &       &    LPIPS $\downarrow$    & 0.13 & 0.13 & 0.14 & 0.09 & 0.08 & 0.15 & 0.11 & 0.15 & 0.12 \\
\cmidrule(lr){2-12} 
    &    \multirow{3}{*}{\makecell{Splat-\\SLAM\cite{sandstrom2024splat}}}    &    \ml{1}{PSNR $\uparrow$}    & 32.25 & \rd34.31 & \rd35.95 & \nd40.81 & \rd40.64 & \rd35.19 & \rd35.03 & \nd37.40 & \rd36.45 \\
    &       &    \ml{1}{SSIM $\uparrow$}    & 0.91 & 0.93 & 0.95 & \fr0.98 & 0.97 & 0.96 & 0.95 & \fr0.98 & 0.95 \\
    &       &    LPIPS $\downarrow$    & 0.10 & \rd0.09 & \nd0.06 & \nd0.05 & \nd0.05 & \nd0.07 & \nd0.06 & \fr0.04 & \nd0.06 \\
\cmidrule(lr){2-12} 
    &    \multirow{3}{*}{\makecell{Ours}}    &    \ml{1}{PSNR $\uparrow$}    & \fr35.48 & \fr36.93 & \fr38.53 & \fr42.28 & \fr43.16 & \fr37.31 & \fr36.99 & \fr38.95 & \fr38.71 \\
    &       &    \ml{1}{SSIM $\uparrow$}    & \rd0.96 & \nd0.97 & \rd0.97 & \fr0.98 & \rd0.98 & \fr0.97 & \fr0.97 & \rd0.97 & \nd0.97 \\
    &       &    LPIPS $\downarrow$    & \fr0.04 & \fr0.04 & \fr0.03 & \fr0.02 & \fr0.03 & \fr0.04 & \fr0.04 & \fr0.04 & \fr0.03 \\
\bottomrule
\end{tabular}}
\end{table}
\begin{table}[h]
\centering
\vspace{-3mm}
\caption{Rendering quality evaluations on ScanNet dataset for rgb and rgbd methods.}\label{tab:scannet_render}
\setlength{\tabcolsep}{3.0pt}          %
\resizebox{0.47\textwidth}{!}{
\begin{tabular}{lclccccccc}
\toprule
& Method & Metric & 0000 & 0059 & 0106 & 0169 & 0181 & 0207 & Avg.\\
\midrule
\multirow{10}{*}{\rotatebox[origin=c]{90}{RGB-D input}}    &    \multirow{3}{*}{\makecell{Point\\-SLAM\cite{sandstrom2023point}}}\,    &    \ml{1}{PSNR $\uparrow$} \,   & 21.30 & 19.48 & 16.80 & 18.53 & 22.27 & 20.56 & 19.82 \\
&       &    \ml{1}{SSIM $\uparrow$}    & 0.81 & 0.77 & 0.68 & 0.69 & 0.82 & 0.75 & 0.75 \\
&       &    LPIPS $\downarrow$    & 0.48 & 0.50 & 0.54 & 0.54 & 0.47 & 0.54 & 0.51 \\
\cmidrule(lr){2-10} 
&    \multirow{3}{*}{\makecell{Splat\\TAM\cite{keetha2024splatam}}}    &    \ml{1}{PSNR $\uparrow$}    & 18.70 & 20.91 & 19.84 & 22.16 & 22.01 & 18.90 & 20.42 \\
&       &    \ml{1}{SSIM $\uparrow$}    & 0.71 & 0.79 & 0.81 & 0.78 & 0.82 & 0.75 & 0.78 \\
&       &    LPIPS $\downarrow$    & 0.48 & 0.32 & 0.32 & 0.34 & 0.42 & 0.41 & 0.38 \\
\cmidrule(lr){2-10} 
&    \multirow{3}{*}{\makecell{Gaussian\\-SLAM\cite{yugay2023gaussian}}}    &    \ml{1}{PSNR $\uparrow$}    & \rd28.54 & \rd26.21 & \rd26.26 & \rd28.60 & \rd27.79 & \rd28.63 & \rd27.67 \\
&       &    \ml{1}{SSIM $\uparrow$}    & \fr0.93 & \fr0.93 & \fr0.93 & \fr0.92 & \fr0.92 & \fr0.91 & \fr0.92 \\
&       &    LPIPS $\downarrow$    & \rd0.27 & \nd0.21 & \rd0.22 & 0.23 & \rd0.28 & \rd0.29 & \rd0.25 \\
\midrule
\multirow{10}{*}{\rotatebox[origin=c]{90}{RGB input}}    &     \multirow{3}{*}{\makecell{GLORIE\\-SLAM\cite{zhang2024glorie}}}    &    \ml{1}{PSNR $\uparrow$}    & 23.42 & 20.66 & 20.41 & 25.23 & 21.28 & 23.68 & 22.45 \\
&       &    \ml{1}{SSIM $\uparrow$}    & \nd0.87 & 0.83 & 0.84 & \nd0.91 & 0.76 & \rd0.85 & 0.84 \\
&       &    LPIPS $\downarrow$    & \nd0.26 & 0.31 & 0.31 & \rd0.21 & 0.44 & \nd0.29 & 0.30 \\
\cmidrule(lr){2-10} 
&    \multirow{3}{*}{\makecell{Splat-\\SLAM\cite{sandstrom2024splat}}}    &    \ml{1}{PSNR $\uparrow$}    & \fr28.68 & \fr27.69 & \nd27.70 & \nd31.14 & \fr31.15 & \fr30.49 & \fr29.48 \\
&       &    \ml{1}{SSIM $\uparrow$}    & 0.83 & \nd0.87 & \rd0.86 & 0.87 & \rd0.84 & 0.84 & \rd0.85 \\
&       &    LPIPS $\downarrow$    & \fr0.19 & \fr0.15 & \fr0.18 & \fr0.15 & \fr0.23 & \fr0.19 & \fr0.18 \\
\cmidrule(lr){2-10} 
&    \multirow{3}{*}{\makecell{Ours}}    &    \ml{1}{PSNR $\uparrow$}    & \nd28.62 & \nd27.22 & \fr28.13 & \fr31.28 & \nd30.37 & \nd30.03 & \nd29.27 \\
&       &    \ml{1}{SSIM $\uparrow$}    & \rd0.85 & \nd0.87 & \nd0.90 & \rd0.90 & \nd0.90 & \nd0.86 & \nd0.88 \\
&       &    LPIPS $\downarrow$    & 0.28 & \rd0.23 & \nd0.21 & \nd0.18 & \nd0.25 & 0.30 & \nd0.24 \\
\bottomrule
\end{tabular}}
\end{table}

Tables~\ref{tab:waymo}, ~\ref{tab:replica_render} and ~\ref{tab:scannet_render} present our rendering quality evaluation results. On the Replica dataset, our system significantly outperforms competing methods, achieving superior PSNR and LPIPS metrics. For the ScanNet dataset, we demonstrate better performance than RGB-D methods while matching the strong baseline of Splat-SLAM~\cite{sandstrom2024splat}. As detailed in our ablation study (Sec.~\ref{sec:ablation}), we could achieve even higher rendering quality by relaxing geometric constraints by normal loss, but this would compromise geometry accuracy. Instead, our system balances the trade-off between geometry and appearance quality.
\begin{figure}[t!]
\centering
\captionsetup{justification=centering}
\includegraphics[width=0.492\textwidth]{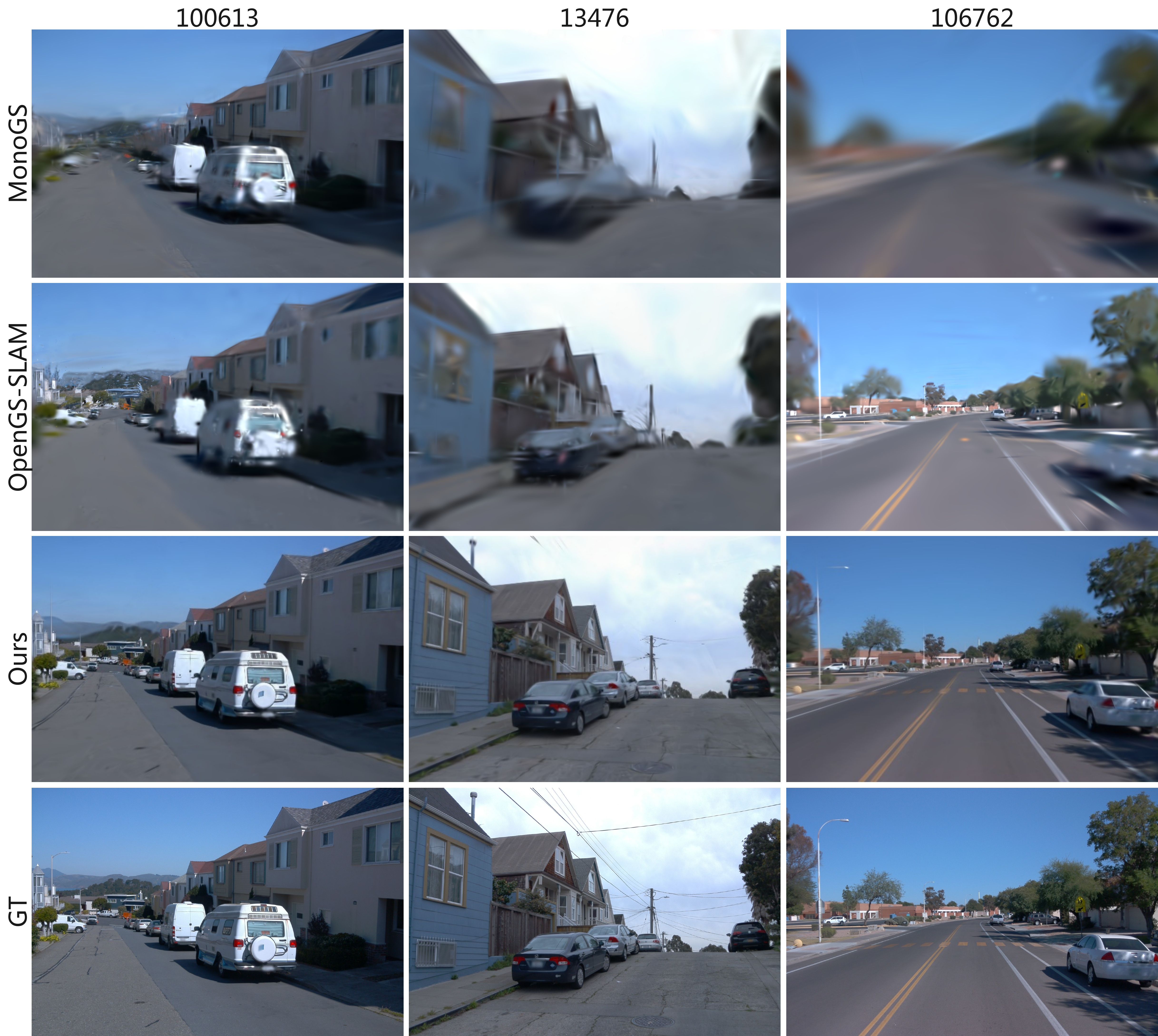}
\caption{Rendering quality comparison on the Waymo Open Dataset in unbounded outdoor scenes. Our method captures finer details of the driving environment, while other methods produce noticeably blurrier results.}\label{fig:comp_waymo}
\end{figure}
On the Waymo Open dataset, Figure~\ref{fig:comp_waymo} shows qualitative comparisons of our rendered RGB images with those from other methods, and Table~\ref{tab:waymo} presents the corresponding quantitative results. Our system achieves significantly higher rendering quality. This enhanced visual fidelity can be attributed to two key factors: firstly our superior tracking accuracy; and secondly the effective integration of depth and normal supervision in our mapping pipeline.

\subsection{Ablation Study}\label{sec:ablation}
\begin{table}[h]
\centering
\vspace{-3mm}
\caption{Depth accuracy of our final rendered depth compared to the prior depth aligned using different strategies, as well as BA depth with and without JDSA assistance, evaluated on the Replica dataset.}\label{tab:ablation_depth}
\setlength{\tabcolsep}{3.0pt}          %
\resizebox{0.48\textwidth}{!}{
\begin{tabular}{lcccccc}
    \toprule
    & \multirow{2}{*}{\makecell{Abs Diff\\$[m]\downarrow$}} & \multirow{2}{*}{\makecell{Abs Rel\\$[\%]\downarrow$}} & \multirow{2}{*}{\makecell{Sq Rel\\$[\%]\downarrow$}} & \multirow{2}{*}{\makecell{RMSE\\$[m]\downarrow$}} & \multirow{2}{*}{\makecell{$\delta<$1.05\\$[\%]\uparrow$}}  & \multirow{2}{*}{\makecell{$\delta<$1.25\\$[\%]\uparrow$}}  \\
    Depth Type & & & & & & \\
    \midrule
    Prior(one scale) & 0.147 & 6.70 & 4.62 & 0.18 & 66.69 & 94.85 \\
    Prior(scale grid) & 0.074 & 3.41 & 0.52 & \rd0.10 & 77.45 & \rd99.66 \\
    BA estimate & \rd0.059 & \rd2.86 & \nd0.37 & \nd0.09 & \rd83.52 & \nd99.74 \\
    BA with JDSA & \nd0.046 & \nd1.99 & \rd0.51 & 0.11 & \nd91.84 & 99.32 \\
    \midrule
    Ours Rendered & \fr0.015 & \fr0.67 & \fr0.10 & \fr0.04 & \fr98.65 & \fr99.81 \\
    \bottomrule
\end{tabular}}
\end{table}

\begin{table}[h]
\centering
\captionsetup{justification=centering}
\ifrevision
\else
\caption{Ablation study on the choice of depth prior models, evaluating rendering quality, reconstruction accuracy, and inference time per frame on the Replica dataset.}\label{tab:depth_prior_ablation}
\fi
\resizebox{0.488\textwidth}{!}{
\begin{tabular}{lccccc}
\toprule
& \makecell[c]{PSNR\\ \,[dB] $\uparrow$}  & \makecell[c]{Acc\\ \,[cm] $\downarrow$} & \makecell[c]{Comp\\ \,[cm] $\downarrow$} & \makecell[c]{Comp Rat\\ \,[\%] $\uparrow$} & \makecell[c]{Runtime\\ \,[ms] $\downarrow$}  \\
\midrule
Ours + Metric3D          & 38.59 & 1.60 & 3.58 & 85.11 & 61 \\
Ours + DA V2             & 38.68 & 1.57 & 3.48 & 85.31 & 32 \\
Ours + ZoeDepth          & 38.60 & 1.64 & 3.60 & 85.00 & 52 \\
Ours + Omnidata          & 38.71 & 1.57 & 3.49 & 85.25 & 6 \\
\bottomrule
\end{tabular}
}
\end{table}

\begin{table}[h]
\centering
\captionsetup{justification=centering}
\ifrevision
\else
\caption{Ablation study on the choice of normal prior models.}\label{tab:normal_prior_ablation}
\fi
\resizebox{0.488\textwidth}{!}{
\begin{tabular}{lccccc}
\toprule
& \makecell[c]{PSNR\\ \,[dB] $\uparrow$}  & \makecell[c]{Acc\\ \,[cm] $\downarrow$} & \makecell[c]{Comp\\ \,[cm] $\downarrow$} & \makecell[c]{Comp Rat\\ \,[\%] $\uparrow$} & \makecell[c]{Runtime\\ \,[ms] $\downarrow$}  \\
\midrule
Ours + EESNU    & 38.60 & 1.70 & 3.64 & 84.61 & 5 \\
Ours + DSINE    & 38.52 & 1.62 & 3.55 & 85.02 & 37 \\
Ours + OmniData & 38.71 & 1.57 & 3.49 & 85.25 & 6 \\
\bottomrule
\end{tabular}
}
\end{table}

\begin{table}[h]
\centering
\caption{Ablation study on the progressive improvement in trajectory accuracy on the Replica dataset, averaged over 8 sequences. From left to right, each stage refines the pose estimation based on the previous stage.}\label{tab:ablation_ate}
\setlength{\tabcolsep}{10.0pt}          %
\resizebox{0.48\textwidth}{!}{
\begin{tabular}{ccccc}
    \toprule
    & \makecell[c]{Online\\Tracking} & \makecell[c]{Online\\PGBA} & \makecell[c]{Offline\\Full BA} & \makecell[c]{Joint Pose Map\\Refinement} \\
    \midrule
    ATE $[cm]\downarrow$ & 0.42 & \rd0.33 & \nd0.32 & \fr0.26 \\
    \bottomrule
\end{tabular}}
\end{table}

\begin{table}[h]
\centering
\caption{Ablation study on the impact of the different proposed modules on the reconstruction performance on the Replica dataset, averaged over 8 sequences.}\label{tab:ablation_module}
\setlength{\tabcolsep}{1.5pt}          %
\resizebox{0.49\textwidth}{!}{
\begin{tabular}{lcccccc}
    \toprule
    & \makecell[c]{PSNR\\ \,[dB] $\uparrow$} & \makecell[c]{SSIM\\$\uparrow$} & \makecell[c]{LPIPS\\$\downarrow$} &\makecell[c]{Acc.\\ \,[cm] $\downarrow$} & \makecell[c]{Comp.\\ \,[cm] $\downarrow$}& \makecell[c]{Comp. Rat\\ \,[\%] $\uparrow$}\\
    \midrule
    w/o grid-based scale align & 37.18 & \rd0.97 & 0.05 & \rd1.68 & \rd3.58 & \rd84.04 \\
    w/o unbiased depth render & \rd38.23 & 0.97 & 0.04 & 1.73 & 3.92 &  81.16 \\
    w/o $\mathcal{L}_{normal}$ & \fr39.09 & \fr0.98 & \fr0.03 &  2.46 & 4.09 &  82.40\\
    w/o joint pose map refine &  37.25 & 0.96 & \rd0.04 & \nd1.61 &  \nd3.55 & \nd84.29\\
    Ours &  \nd38.71 & \nd0.97 & \nd0.03 &  \fr1.57 &  \fr3.49 &  \fr85.25\\
    \bottomrule
\end{tabular}}
\end{table}

\textbf{Monocular Prior Integration} We first evaluate the depth estimation accuracy improved by our different modules on the Replica dataset, as depth accuracy is crucial for scene reconstruction quality. Table~\ref{tab:ablation_depth} quantifies the effectiveness of different approaches. Comparing Prior (one scale) with Prior (scale grid) demonstrates that our grid-based scale alignment significantly outperforms the single-scale alignment from HI-SLAM~\cite{zhang2024hi}, better addressing inherent scale distortions in monocular depth priors. Further analysis compares two depth estimation approaches: BA estimate (using BA alone) and BA with JDSA (using interleaved BA and JDSA optimization). The results confirm that incorporating JDSA with alternating optimization outperforms BA alone, validating our depth prior integration strategy. The final rendered depth from our Gaussian map achieves the highest accuracy, validating the effectiveness of our complete pipeline.

We further investigate the incorporation of alternative depth \revadd{and normal} prior predictors. For depth priors, including Metric3D~\cite{yin2023metric3d}\revadd{, ZoeDepth~\cite{bhat2023zoedepth},} and Depth Anything (DA) V2~\cite{yang2024depth}, we present results in Table~\ref{tab:depth_prior_ablation}. \revadd{Notably, ZoeDepth and DA V2 were not trained on any Replica images, yet still achieve comparable performance, indicating that our method does not rely on indirect biases from the training data. For normal priors, we evaluate EESNU~\cite{bae2021estimating} and DSINE~\cite{bae2024rethinking}, as shown in Table~\ref{tab:normal_prior_ablation}.} The results demonstrate that our method remains compatible with these alternatives. Nevertheless, our chosen prior, OmniData, offers a more efficient trade-off between performance and computational cost. \revadd{This choice also aligns with the baseline methods, which likewise employ OmniData as the geometry prior.}
\begin{figure}[t!]
\centering
\includegraphics[width=0.48\textwidth]{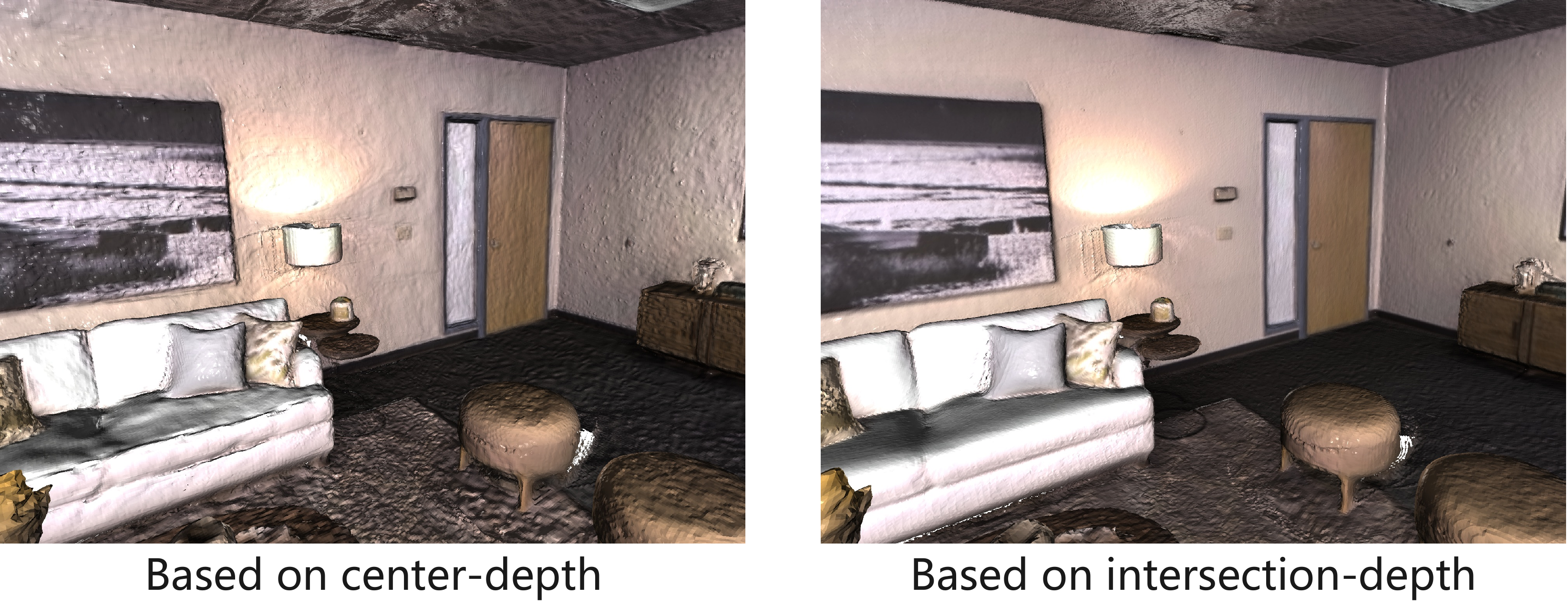}
\vspace*{-1mm}
\caption{Reconstruction quality comparison on the \textit{Room0} scene of the Replica dataset, using rendered depth based on the depth at the Gaussian center versus our approach, which uses the depth at the ray-Gaussian intersection.}
\label{fig:depth-compare}
\end{figure}

\textbf{Trajectory Accuracy} Table~\ref{tab:ablation_ate} demonstrates the progressive improvement in pose estimation accuracy on the Replica dataset through our system pipeline. Starting with initial estimates from the online tracking module, accuracy is first enhanced through online PGBA based loop closing. A subsequent full BA further refines these results, with the final joint pose and 3DGS map refinement achieving the highest trajectory accuracy. This systematic improvement across stages validates the effectiveness of our hierarchical optimization approach.

\textbf{Component Analysis} To evaluate key design components, we conduct ablation studies by removing individual components. Table~\ref{tab:ablation_module} confirms each module's contribution to system performance. The grid-based scale alignment proves crucial, as its removal significantly degrades reconstruction accuracy. Similarly, the unbiased depth rendering enhances both rendering quality and geometric accuracy, with qualitative comparison shown in Fig.~\ref{fig:depth-compare}. While the normal loss slightly affects appearance metrics, it substantially improves geometry quality. The final joint pose and map refinement further enhances accuracy through improved global consistency.

\subsection{Performance Analysis}\label{sec:runtimemapsize}
\begin{figure}[!htb]
\includegraphics[width=0.49\textwidth,left]{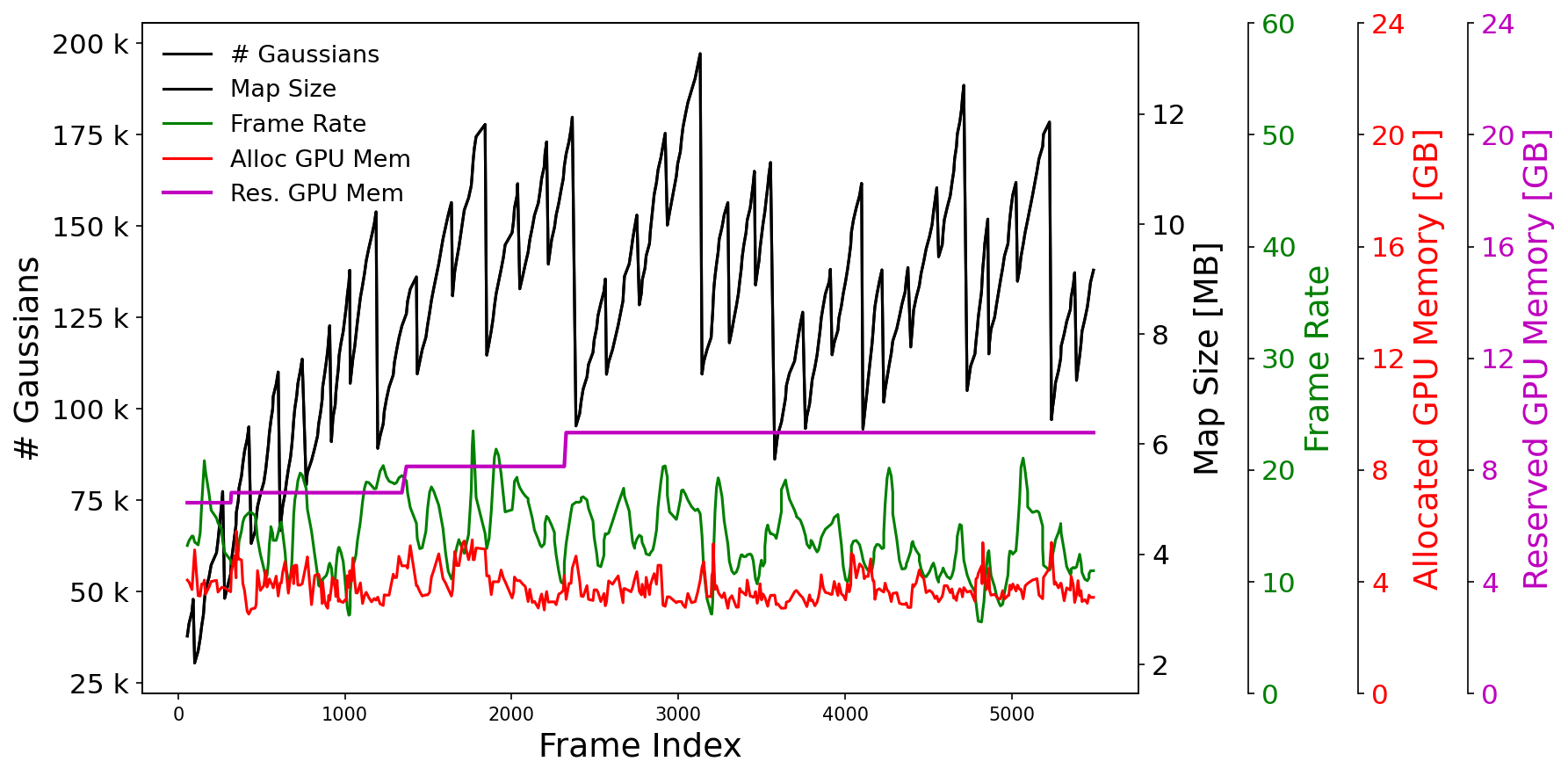}
\vspace*{-3mm}
\ifrevision
\else
\caption{Evolution of map size, GPU memory usage, and system speed over frame index for the \textit{Scene0000} sequence from ScanNet, evaluated on an RTX 4090 GPU. The number of Gaussians grows as new areas are explored and stabilizes with the pruning strategy. System speed denotes the processing frame rate for online tracking, loop closing, and mapping. Allocated and reserved GPU memory usage are both reported to provide resource analysis.}\label{fig:mapsize}
\fi
\end{figure}
Our system achieves real-time performance with the online tracking, loop closing and mapping operating at 22 frames per second (FPS) on the Replica dataset, and takes only 12 seconds for offline refinement. On the ScanNet dataset, despite more rapid camera movements, it maintains robust performance online above 10 FPS, with offline refinement taking a few minutes due to a higher number of optimization iterations. Fig.~\ref{fig:mapsize} illustrates the map size evolution\revadd{, GPU memory consumption, and processing speed} for sequence \textit{Scene0000}. While the Gaussian count initially grows during new area exploration, our efficient map pruning strategy stabilizes the map size by eliminating redundant Gaussians in revisited regions.\revadd{ The memory analysis reveals that both allocated and reserved GPU memory usage remain stable throughout the sequence, with periodic fluctuations corresponding to pruning cycles. Reserved memory gradually increases as additional buffers are allocated to store keyframe states. The processing frame rate decreases during rapid motion due to more frequent keyframe insertion, and increases during stable motion, remaining consistent across different phases.} This demonstrates our system's scalability and efficiency in managing large-scale environments\revadd{ while maintaining predictable resource consumption}.

\begin{figure*}[t!]
\centering
\includegraphics[width=0.98\textwidth]{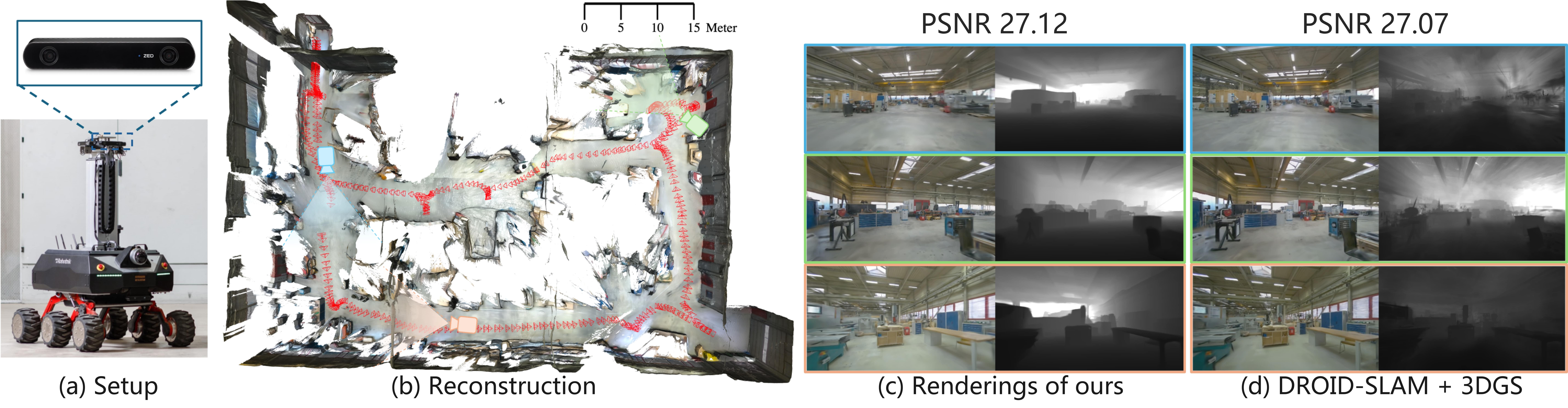}
\vspace*{-2mm}
\caption{Evaluation on self-collected data with the robot navigating through a large factory hall. Our robot (a) is equipped with stereo cameras. Our system operates solely on the monocular input from the left camera and reconstructs the scene along with the estimated camera trajectory (b). The rendered color and depth maps (c) achieve comparable visual quality while exhibiting significantly better geometry and free of floaters\revdel{ that are present in the baseline DROID-SLAM + 3DGS (d).}\revadd{. In contrast, the baseline DROID-SLAM + 3DGS (d) suffers from geometric artifacts and floating points. We refer readers to our supplementary video, which showcases novel view renderings on out-of-sequence viewpoints.}}\label{fig:robot}
\vspace*{-5mm}
\end{figure*}

\subsection{Evaluation on Self-Collected Robot Data}\label{sec:robot}
\begin{figure}[t!]
\includegraphics[width=0.49\textwidth]{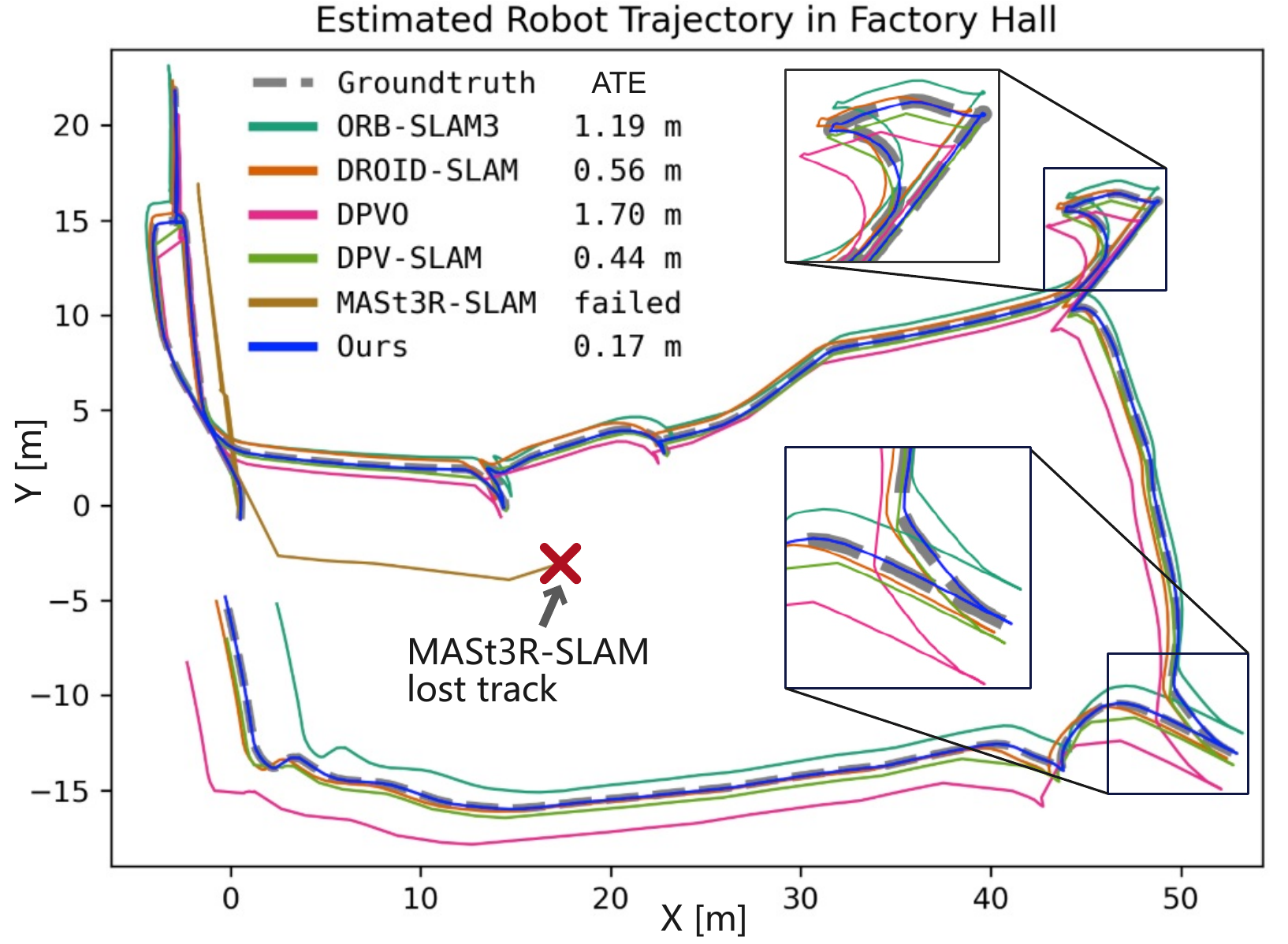}
\vspace{-6mm}
\ifrevision
\else
\caption{Qualitative comparison of the estimated robot trajectory in the factory hall.}\label{fig:robot-traj}
\fi
\end{figure}   
We evaluate our system on a self-collected dataset captured by our robotic platform. Figure~\ref{fig:robot} shows the platform and the results in a large factory hall. The robot is equipped with stereo cameras. However, only the monocular input from the left camera is used for this experiment. The recorded sequence has intotal 4073 frames with a duration of 6 minutes and 52 seconds.
We compare our rendering results with the baseline method DROID-SLAM + 3DGS, where the estimated camera poses and point cloud from DROID-SLAM serve as input and used to initialize the Gaussians. For our system, it completes online stage including tracking, loop closuing, and mapping in 5 minutes and 20 seconds, followed by 3 minutes and 32 seconds for the offline refinement stage. In contrast, the baseline requires 4 minutes and 24 seconds for DROID-SLAM and an additional 15 minutes and 2 seconds for 3DGS mapping. Overall, our method takes only about half the runtime of the baseline while achieving comparable rendering quality and significantly better geometry, without the floaters observed in the baseline results.\revadd{ Furthermore, we evaluate the trajectory accuracy using the centimeter-accurate photogrammetric reference described in~\cite{ress20253d}, as shown in Fig.~\ref{fig:robot-traj}. While MASt3R-SLAM~\cite{murai2024mast3r} struggles to track camera poses due to poor generalizability to the new environment, our system maintains stable tracking and successfully detects loop closures with the highest accuracy.}

\subsection{Extention to Semantic Reconstruction}\label{sec:semantic}
\begin{figure}[!htb]
\includegraphics[width=0.49\textwidth,left]{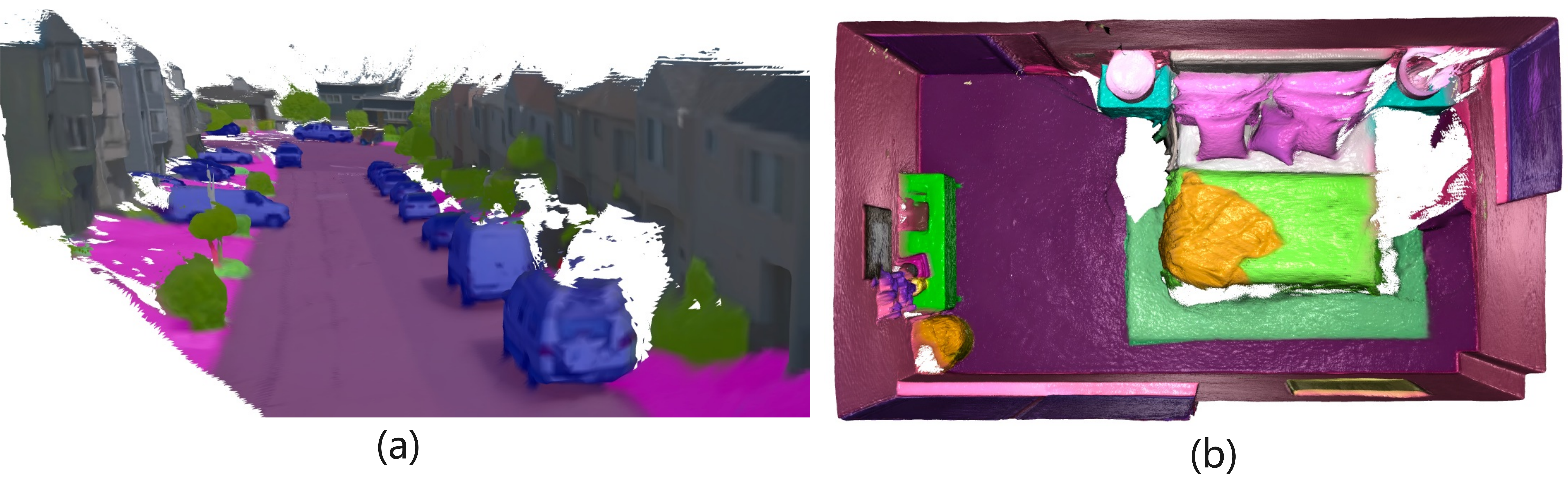}
\vspace*{-5mm}
\caption{Semantic reconstruction results: (a) outdoor scene from \textit{100613} of Waymo Open dataset, (b) indoor scene from \textit{Room1} of the Replica dataset.}\label{fig:semantics}
\end{figure}
As an extension, we demonstrate the capability of our system for semantic scene reconstruction by incorporating 2D semantic information into the 3DGS representation. Each Gaussian primitive is augmented with semantic color channels in addition to its existing geometric and appearance attributes. These semantic channels can be efficiently rasterized to the image plane alongside color and depth, enabling simultaneous semantic colorization of the reconstructed scene. For semantic optimization, we maintain the same pipeline structure as our depth and pose optimization framework, with an additional L1 semantic RGB loss term that measures the absolute difference between rendered and ground truth semantic color maps. Following the evaluation protocol of~\cite{haghighi2023neural}, we assess our semantic reconstruction performance on the Replica dataset using mean Intersection over Union (mIoU) as the primary metric. We evaluate on four standard sequences to enable direct comparison with existing baseline methods.

\begin{table}[h]
\centering
\ifrevision
\else
\caption{Semantic reconstruction results evaluated by mIoU metric on 4 sequences of the Replica dataset.}\label{tab:semantic_replica}
\fi
\resizebox{0.47\textwidth}{!}{
\begin{tabular}{lcccccc}
\toprule
& Method & ro-0 & ro-1 & ro-2 & of-0 & Avg.mIoU[\%]$\uparrow$\\
\midrule
\multirow{4}{*}{\rotatebox[origin=c]{90}{RGB-D}}   &   NIDS-SLAM\cite{haghighi2023neural}   & 82.45 & 84.08 & 76.99 & 85.94 & 82.37 \\
      &   DNS-SLAM\cite{li2024dns}   & 88.32 & 84.90 & 81.20 & 84.66 & 84.77 \\
      &   SNI-SLAM\cite{zhu2024sni}   & 88.42 & 87.43 & 86.16 & 87.63 & 87.41 \\
      &   SGS-SLAM\cite{li2024sgs}    & \nd{92.95} & \nd{92.91} & \nd{92.10} & \nd{92.90} & \nd{92.72} \\
      &   Hier-SLAM\cite{li2025hier}    & \fr{95.25} & \fr{95.81} & \fr{95.73} & \fr{95.52} & \fr{95.58} \\
\midrule
\multirow{2}{*}{\rotatebox[origin=c]{90}{RGB}}   &   HI-SLAM\cite{zhang2024hi}  & 74.93 & 79.55 & 80.90 & 71.53 & 76.72 \\
      &   Ours  & \rd{90.27} & \rd{92.80} & \rd{91.11} & \rd{92.45} & \rd{91.65} \\
\bottomrule
\end{tabular}}
\end{table}
Tab.~\ref{tab:semantic_replica} presents quantitative results comparing our approach against recent RGB-D and RGB-only semantic SLAM methods\revadd{, including concurrent work Hier-SLAM~\cite{li2025hier}}. Our system achieves competitive performance compared to state-of-the-art RGB-D methods, while significantly outperforming the RGB-only baseline HI-SLAM~\cite{zhang2024hi} by a margin of 14.93\%. This substantial improvement can be attributed to two key factors: (1) our more accurate geometry reconstruction provides better surface boundaries for semantic label assignment, and (2) the explicit 3DGS representation allows for sharper semantic boundaries compared to implicit NeRF-based approaches that often struggle with object delineation in complex scenes or regions with small objects. Fig.~\ref{fig:semantics} demonstrates qualitative results: indoor scene from the Replica dataset with ground truth semantic labels, and outdoor driving scene from the Waymo Open dataset using Mask2Former~\cite{cheng2021mask2former} predictions as semantic inputs.

\section{Conclusion}
This article presents HI-SLAM2, a novel monocular SLAM system that achieves fast and accurate dense 3D scene reconstruction through four complementary modules. The online tracking module enhances depth and pose estimation by integrating depth priors with grid-based scale alignment, while parallel PGBA in the online loop closing module corrects pose and scale drift. Our mapping approach leverages 3D Gaussian splatting for compact scene representation, continuously refined during SLAM tracking. We enhance geometric consistency through monocular normal priors and unbiased ray-Gaussian intersection depth for splat-based rasterization. During offline refinement, we achieve high-fidelity reconstruction by incorporating exposure compensation and performing joint optimization of camera poses, 3DGS map, and exposure parameters. Extensive evaluations on challenging datasets demonstrate that HI-SLAM2 outperforms state-of-the-art methods in accuracy and completeness while maintaining superior runtime performance. Our system achieves high-quality geometry and appearance reconstruction without the typical trade-offs observed in other methods. \\
\textbf{Limitations:} Our system has three main limitations: First, the current proximity-based loop closure detection shows limited robustness in the ETH3D dataset when encountering view occlusions and textureless regions, suggesting the need for learned feature-based place recognition. Second, in city-scale scenes, mapping quality can degrade due to limited optimization budget, indicating the need for submap optimization strategies. Third, the system assumes static environments. Incorporating dynamic object detection and tracking, along with motion segmentation, would enable robust operation in dynamic environments.

\section*{Acknowledgment}
This work is partially supported by the Deutsche Forschungsgemeinschaft (DFG, German Research Foundation) under Germany's Excellence Strategy - EXC 2120/1 - project number 390831618.
Qing Cheng is supported by the DAAD program Konrad Zuse Schools of Excellence in Artificial Intelligence, and the Federal Ministry of Research, Technology and Space.

\bibliographystyle{IEEEtran}
\bibliography{IEEEabrv,root}

\end{document}